\documentstyle{esub2acm}
\newtheorem{theorem}{Theorem}[section]
\newtheorem{lemma}[theorem]{Lemma}
\newtheorem{corollary}[theorem]{Corollary}
\newtheorem{proposition}[theorem]{Proposition}
\newtheorem{definition}[theorem]{Definition}

\newtheorem{example}[theorem]{Example}
\newtheorem{remark}[theorem]{Remark}
\newcommand{\thm}{\begin{theorem}}
\newcommand{\lem}{\begin{lemma}}
\newcommand{\pro}{\begin{proposition}}
\newcommand{\dfn}{\begin{definition}}
\newcommand{\rem}{\begin{remark}}
\newcommand{\xam}{\begin{example}}
\newcommand{\cor}{\begin{corollary}}
\newcommand{\prf}{\proof}
\newcommand{\ethm}{\end{theorem}}
\newcommand{\elem}{\end{lemma}}
\newcommand{\epro}{\end{proposition}}
\newcommand{\edfn}{\bbox\end{definition}}
\newcommand{\erem}{\bbox\end{remark}}
\newcommand{\exam}{\bbox\end{example}}
\newcommand{\ecor}{\end{corollary}}
\newcommand{\eprf}{\endproof}
\newcommand{\beqn}{\begin{equation}}
\newcommand{\eeqn}{\end{equation}}
\newcommand{\bbox}{\vrule height7pt width4pt depth1pt}

\newcommand{\commentout}[1]{}
\newcommand{\denselist}{\itemsep 3pt}

\newcommand{\eg}{e.g.,~}
\newcommand{\ie}{i.e.,~}
\newcommand{\resp}{resp.\ }
\newcommand{\respc}{resp.,\ }
\newcommand{\A}{{\cal A}}

\newcommand{\F}{{\cal F}}

\renewcommand{\P}{{\cal P}}

\newcommand{\IN}{\mbox{$I\!\!N$}}

\newcommand{\false}{\mbox{{\it false}}}
\newcommand{\true}{\mbox{{\it true}}}

\newcommand{\sat}{\models}
\newcommand{\rimp}{\Rightarrow}
\newcommand{\dimp}{\Leftrightarrow}
\newcommand{\bor}{\bigvee}
\newcommand{\band}{\bigwedge}
\newcommand{\union}{\cup}
\newcommand{\inter}{\cap}

\renewcommand{\L}{{\cal L}}
\renewcommand{\S}{{\cal S}}

\newcommand{\Cond}{\mbox{\boldmath$\rightarrow$\unboldmath}}
\newcommand{\nCond}{\mbox{\boldmath$\not\rightarrow$\unboldmath}}

\newcommand{\True}{\mbox{\it true}}
\newcommand{\False}{\mbox{\it false}}

\newcommand{\intension}[1]{[\![ #1 ]\!]}

\newcommand{\Pl}{\mbox{\rm Pl\/}}
\newcommand{\PL}{\mbox{\it PL}}
\newcommand{\PBox}{N}
\newcommand{\bottom}{\perp}
\newcommand{\sPL}{{\mbox{\scriptsize\it PL}}}

\newcommand{\Poss}{\mbox{Poss}}
\newcommand{\sPoss}{\mbox{\scriptsize\it Poss}}

\newcommand{\LCond}{\L^{C}}
\newcommand{\SysP}{system~{\bf P}}

\newcommand{\Union}{\bigcup}

\newcommand{\Tref}[1]{Theorem~\ref{#1}}
\newcommand{\Lref}[1]{Lemma~\ref{#1}}
\newcommand{\Pref}[1]{Proposition~\ref{#1}}
\newcommand{\Cref}[1]{Corollary~\ref{#1}}

\newcommand{\Sref}[1]{Section~\ref{#1}}

\newcommand{\Sub}{\mbox{{\em Sub}}}

\newenvironment{RETHM}[2]{\trivlist \item[\hskip 10pt\hskip\labelsep{\sc #1\hskip 5pt\relax\ref{#2}.}]\it}{\endtrivlist}
\newcommand{\rethm}[1]{\begin{RETHM}{Theorem}{#1}}
\newcommand{\repro}[1]{\begin{RETHM}{Proposition}{#1}}
\newcommand{\relem}[1]{\begin{RETHM}{Lemma}{#1}}
\newcommand{\recor}[1]{\begin{RETHM}{Corollary}{#1}}

\newcommand{\erethm}{\end{RETHM}}
\newcommand{\erepro}{\end{RETHM}}
\newcommand{\erelem}{\end{RETHM}}
\newcommand{\erecor}{\end{RETHM}}

\newcommand{\BelF}{\mbox{\it Bel}}

\newcommand{\PS}{\mbox{\it PS}}
\renewcommand{\Omega}{W}
\newcommand{\SysC}{system~{\bf C}}
\newcommand{\vdashL}{\vdash_{\L}}
\newcommand{\vdashp}{\vdash_{\mbox{\scriptsize\bf P}}}
\newcommand{\vdashpp}{\vdash_{\mbox{\scriptsize\bf P}'}}
\newcommand{\vdashc}{\vdash_{\mbox{\scriptsize\bf C}}}

\newcommand{\satp}{\sat_{\mbox{\scriptsize p}}}
\newcommand{\satr}{\sat_{\mbox{\scriptsize r}}}
\newcommand{\sate}{\sat_\epsilon}
\newcommand{\satPoss}{\sat_{\sPoss}}
\newcommand{\satkappa}{\sat_\kappa}
\newcommand{\satPL}{\sat_{PL}}

\newcommand{\LDef}{\L_{\mbox{\scriptsize def}}}
\newcommand{\Ele}{\unlhd}
\newcommand{\Elt}{\lhd}

\newcommand{\sQPL}{QPL}

\newcommand{\IntensionV}[1]{[#1]_{w_V}}

\newcommand{\Plass}{\P}
\newcommand{\PLclass}{\S}

\renewcommand{\Omega}{W}

\newcommand{\Bird}{\mbox{\it Bird\/}}

\newcommand{\Fly}{\mbox{\it Fly\/}}

\begin{document}
\title{Plausibility Measures and Default Reasoning}
\author{
Nir Friedman\\
University of California, Berkeley
\and
Joseph Y.\ Halpern\\
Cornell University
}

\begin{abstract}
We introduce a new approach to modeling uncertainty based on {\em
    plausibility measures}.  This approach is easily seen to
    generalize other approaches to modeling uncertainty, such as
    probability measures, belief functions, and possibility measures.
    We focus on one application of plausibility measures in this
    paper: default reasoning.  In recent years, a number of different
    semantics for defaults have been proposed, such as preferential
    structures, $\epsilon$-semantics, possibilistic structures, and
    $\kappa$-rankings, that have been shown to be characterized by the
    same set of axioms, known as the KLM properties.
    While this was viewed as a surprise, we
    show here that it is almost inevitable.  In the framework of
    plausibility measures, we can give a necessary condition for the
    KLM axioms to be sound, and an additional condition necessary and
    sufficient to ensure that the KLM axioms are complete.  This
    additional condition is so weak that it is almost always met
    whenever the axioms are sound. In particular, it is easily seen to
    hold for all the proposals made in the literature.
\end{abstract}

\category{F.4.1}{Mathematical Logic and Formal Languages}{Mathematical
    Logic}
\category{I.2.3}{Artificial Intelligence}{Deduction and Theorem
Proving}[Nonmonotonic reasoning and belief revision] 
\category{I.2.4}{Artificial Intelligence}{Knowledge Representation
    Formalisms and Methods}[Representation Languages]
\terms{Theory}
\keywords{Conditional Logic, Default Reasoning, $\epsilon$-Semantics,
$\kappa$-rankings, Plausibility Measures, Possibility Measures,
Preferential Orderings
Nonmonotonic Inference.} 
\begin{bottomstuff}
Some of this work was done while both authors were at the IBM Almaden
Research Center. The first author was also at Stanford while much of
the work was done.  IBM and Stanford's support are gratefully
acknowledged.  The work was also supported in part by the Air Force
Office of Scientific Research (AFSC), under Contract F49620-91-C-0080
and grant F94620-96-1-0323 and by NSF under grants IRI-95-03109 and
IRI-96-25901. The first author was also supported in part by
an IBM Graduate Fellowship and by the Rockwell Science Center.
A preliminary version of this paper
appears in {\em Proceedings,
Thirteenth
National Conference on
                     Artificial Intelligence\/}, 1996, pp.~1297--1304.
This version of the paper is almost identical to one that will appear in
{\em JACM}.

\begin{authinfo}
\name{Nir Friedman}
\address{
Computer Science Division, 387 Soda Hall,
University of California, Berkeley, CA 94720.
email: {\tt nir@cs.berkeley.edu};
url: {\tt http://www.cs.berkeley.edu/$\sim$nir}
}
\name{Joseph Y.~Halpern}
\address{
Computer Science Department,
Cornell University, Ithaca, NY 14853.\\
email: {\tt halpern@cs.cornell.edu};
url: {\tt http://www.cs.cornell.edu/home/halpern}
}
\end{authinfo}
\end{bottomstuff}
\markboth{Nir Friedman and Joseph Y.~Halpern}{Plausibility Measures
and Default Reasoning}
\maketitle

\section{Introduction}

We must reason and act in an uncertain world.  There may be
    uncertainty about the state of the world, uncertainty about the
    effects of our actions, and uncertainty about other agents'
    actions. The standard approach to modeling uncertainty is
    probability theory. In recent years, researchers, motivated by
    varying concerns including a dissatisfaction with some of the
    axioms of probability and a desire to represent information more
    qualitatively, have introduced a number of generalizations and
    alternatives to probability, such as Dempster-Shafer belief
    functions \cite{Shaf} and possibility theory \cite{DuboisPrade88}.
    Our aim here is to introduce what is perhaps the most general
    approach possible,
which uses what we call {\em plausibility measures}.  A
    plausibility measure associates with a set a {\em plausibility},
    which is just an element in a partially ordered space.  The only
    property that we impose
    is that the plausibility of a set be at
    least as large as the plausibility of any of its subsets.  Every
    systematic approach for dealing with uncertainty that we are aware
    of can be viewed as a plausibility measure.  Given how little
    structure we have imposed on plausibility measures, this is
    perhaps not surprising.  Nevertheless, as we hope to demonstrate
    in this and other papers, plausibility measures provide us with a
    natural setting in which to examine various approaches to
    reasoning about uncertainty.

The focus of this paper is (propositional) default reasoning.  There
    have been many approaches to default reasoning proposed in the
    literature (see \cite{Ginsberg87,HLIAILP:v3} for overviews).  The
    recent literature has been guided by a collection of axioms that
    have been called the {\em KLM properties\/} (since they were
    discussed by Kraus, Lehmann, and Magidor \citeyear{KLM}), and many
of the recent approaches to
    default reasoning, including {\em preferential structures\/}
    \cite{KLM,Shoham87}, {\em $\epsilon$-semantics\/}
    \cite{Adams:75,Geffner92,Pearl90}, {\em possibilistic structures\/}
    \cite{DuboisPrade:Defaults91}, and {\em $\kappa$-rankings\/}
    \cite{Goldszmidt92,spohn:88}, have been shown to be characterized
    by these properties.  This has been viewed as somewhat surprising,
    since these approaches seem to capture quite different intuitions.
    As Pearl \citeyear{Pearl90} said of the equivalence between
    $\epsilon$-semantics and preferential structures, ``It is
    remarkable that two totally different interpretations of defaults
    yield identical sets of conclusions and identical sets of
    reasoning machinery.''  As we shall show in this paper,
    plausibility measures help us understand why this should be so.

In fact, we show much more.  All of these approaches can be understood
    as giving semantics to defaults by considering a class $\P$ of
    structures (preferential structures, possibilistic structures,
    etc.).  A default $d$ is then said to follow from a knowledge base
    $\Delta$ of defaults if all structures in $\P$ that satisfy
    $\Delta$ also satisfy $d$.  We define a notion of {\em qualitative
    plausibility measure}, and show that the KLM properties are sound
    in a plausibility structure if and only if it is qualitative.
    Moreover, as long as a class $\P$ of plausibility structures
    satisfies a minimal richness condition, we show that the KLM
    properties will completely characterize default reasoning in $\P$.
    We then show that when we map preferential structures (or
    possibilistic structures or any of the other structures considered
    in the literature on defaults) into plausibility structures, we
    get a class of qualitative structures that is easily seen to
    satisfy the richness condition.  This explains why the KLM
    axioms characterize default reasoning in all these frameworks.
    Far from being surprising that the KLM axioms are complete in all
    these cases, it is almost inevitable.

The KLM properties have been viewed as the ``conservative core'' of
    default reasoning \cite{Pearl90}, and much recent effort has been
    devoted to finding principled methods of going beyond KLM.
Our
    result
suggests that it will be difficult to find an interesting approach that
gives semantics to defaults
    with respect to a collection $\P$ of structures
and goes beyond
beyond KLM.  This result thus provides added
    insight into and justification for approaches such as those of
    \cite{BGHKnonmon,geffner:92,Goldszmidt92,GMPfull,LehmannMagidor,Pearl90B}
    that, roughly speaking, say $d$ follows from $\Delta$ if $d$ is
    true in a particular structure $P_\Delta \in \P$ that satisfies
    $\Delta$ (rather than requiring that $d$ be true in all structures
in $\P$ that satisfy   $\Delta$).

This paper is organized as follows. In Section~\ref{plausibility}, we
introduce plausibility measures and show how they generalize various
other proposals for capturing uncertainty.  In Section~\ref{default},
we review the KLM properties and various approaches to default
reasoning that are characterized by these properties.  In
Section~\ref{qualitative plausibility}, we show how the various
notions of default reasoning considered in the literature can all be
viewed as instances of plausibility measures.  In Section~\ref{axioms},
we prove our main results: we define qualitative plausibility
structures, show that the KLM properties are sound in a structure if
and only if it is qualitative, and provide a weak richness condition
that is necessary and sufficient for them to be complete.
In Sections~\ref{sec:epsilon},~\ref{entrenchment},
and~\ref{conditionallogic}, we expand on three independent topics
related to our results.
In Section~\ref{sec:epsilon}, we show that qualitative plausibility
measures are more expressive than previous semantics considered in the
literature.
In Section~\ref{entrenchment}, we consider related work, focusing
on the relationship between our approach to plausibility and
{\em epistemic entrenchment} \cite{MakGar,MakGar:Nonmon,Grove}.
In
Section~\ref{conditionallogic}, we discuss how plausibility measures
can be used to give semantics to a full logic of conditionals, and
compare this with the more traditional approach \cite{Lewis73}.
We conclude in Section~\ref{discussion}
    with a discussion of other potential applications of plausibility
    measures.

\section{Plausibility Measures}
\label{plausibility}
\label{PLAUSIBILITY}
A probability space is a tuple $(\Omega,\F,\mu)$, where $\Omega$ is a
    set of worlds, $\F$ is an algebra of {\em measurable\/} subsets of
    $\Omega$ (that is, a set of subsets closed under union and
    complementation to which we assign probability), and $\mu$ is a
    {\em probability measure}, that is, a function mapping each set in
    $\F$ to a number in $[0,1]$ satisfying the well-known Kolmogorov
    axioms ($\mu(\emptyset) = 0$, $\mu(\Omega) = 1$, and $\mu(A \union
    B) = \mu(A) + \mu(B)$ if $A$ and $B$ are disjoint).%
\footnote{Frequently it is also  assumed that $\mu$ satisfies
    {\em countable additivity\/}, \ie if $A_i$, $i > 0$, are pairwise
    disjoint, then $\mu(\Union_{i} A_i) = \sum_i \mu(A_i)$.}

A plausibility space is a direct generalization of a probability
    space.  We simply replace the probability measure $\mu$ by a {\em
    plausibility measure\/} $\Pl$ that, rather than mapping sets in
    $\F$ to numbers in $[0,1]$, maps them to elements in some
    arbitrary partially ordered set.  We read $\Pl(A)$ as ``the
    plausibility of set $A$''.  If $\Pl(A) \le \Pl(B)$, then $B$ is at
    least as plausible as $A$.  Formally, a {\em plausibility space\/}
    is a tuple $S = (\Omega,\F, \Pl)$, where $\Omega$ is a set of
    worlds, $\F$ is an algebra of subsets of $\Omega$, and $\Pl$ maps
    the sets in $\F$ to some set $D$, partially ordered by a relation
    $\le_D$ (so that $\le_D$ is reflexive, transitive, and
    anti-symmetric).  We assume that $D$ is {\em pointed\/}, that is,
    it contains two special elements $\top_D$ and $\bottom_D$ such
    that $\bottom_D \le_D d \le_D \top_D$ for all $d \in D$; we
    further assume that $\Pl(\Omega) = \top_D$ and $\Pl(\emptyset) =
    \bottom_D$.  The only other assumption we make is
\begin{itemize}
 \item[\bf A1.] If $A \subseteq B$, then $\Pl(A) \le_D \Pl(B)$.
\end{itemize}
Thus, a set must be at least as plausible as any of its subsets.

Some brief remarks on the definition: We have deliberately suppressed
    the domain $D$ from the tuple $S$, since the choice of $D$ is not
    significant in this paper.  All that matters is the ordering
    induced by $\le_D$ on the subsets in $\F$.%
\footnote{In dealing with
    {\em conditional\/} plausibility, the domain $D$ plays a more significant
    role
\cite{FrH7}.}
As usual, we define the ordering $<_D$ by taking
    $d_1 <_D d_2$ if $d_1 \le_D d_2$ and $d_1 \ne d_2$.  We omit the
    subscript $D$ from $\le_D$, $<_D$, $\top_D$, and $\bot_D$ whenever
    it is clear from context.
We also frequently omit the $\F$ when describing a plausibility space
when its role is not that significant, and just
denote a plausibility space as a pair $(\Omega,\Pl)$ rather than
$(\Omega,\F,\Pl)$.

Clearly plausibility spaces generalize probability spaces.  We now
    briefly discuss a few other notions of uncertainty that they
    generalize:
\begin{itemize}
\item A {\em belief function\/} $\BelF$ on $W$ is a function
    $\BelF: 2^{W} \rightarrow [0,1]$ satisfying certain axioms
    \cite{Shaf}.  These axioms certainly imply property A1, so a
    belief function is a plausibility measure.

\item A {\em fuzzy measure\/} (or a {\em Sugeno measure\/}) $f$ on
    $\Omega$ \cite{WangKlir} is a function $f:2^{\Omega}\mapsto
    [0,1]$, that satisfies A1 and some continuity constraints.
A {\em possibility measure\/} \cite{DuboisPrade88} $\Poss$ is a fuzzy
measure such that $\Poss(W) = 1$, $\Poss(\emptyset) = 0$, and
$\Poss(A) = \sup_{w \in A}(\Poss(\{w\})$.

\item An {\em ordinal ranking\/} (or {\em $\kappa$-ranking\/}) $\kappa$
on
    $\Omega$ (as defined by \cite{Goldszmidt92}, based on ideas that
    go back to \cite{spohn:88}) is a function
mapping subsets of $\Omega$ to $\IN^* = \IN \union \{\infty\}$
    such
    that $\kappa(\Omega) = 0$, $\kappa(\emptyset) = \infty$, and
    $\kappa(A) = \min_{w\in A}(\kappa(\{w\}))$. Intuitively, an
    ordinal ranking assigns a degree of surprise to each subset of
    worlds in $\Omega$, where $0$ means unsurprising and higher
    numbers denote greater surprise. It is easy to see that if
    $\kappa$ is a ranking on $\Omega$, then $(\Omega, \kappa)$
    is a plausibility space, where $x \le_{\IN^*} y$ if and only if $y
    \le x$ under the usual ordering on the ordinals.

\item A {\em preference ordering\/} on $W$ is a strict partial order
    $\prec$ over $W$ \cite{KLM,Shoham87}. Intuitively, $w \prec w'$
    holds if $w$ is {\em
    preferred\/} to $w'$.%
\footnote{We follow the standard notation for preference here
    \cite{Lewis73,KLM}, which uses the (perhaps confusing) convention
    of placing the more likely
(or less abnormal)
world on the left of the $\prec$
    operator.}
Preference orders have been used to provide
    semantics for {\em default\/} (\ie conditional) statements.
    In \Sref{qualitative plausibility} we show how to map preference
    orders on $\Omega$ to
    plausibility measures on $W$ in a way that preserves the
    ordering on worlds as well as the truth
    values of defaults.

\item A {\em parameterized probability distribution \/} (PPD) on $W$ is a
    sequence $ \{\Pr_i : i \ge 0\}$ of
    probability measures over $W$. Such structures provide semantics
    for defaults in {\em $\epsilon$-semantics\/}
    \cite{Pearl90,GMPfull}.  In \Sref{qualitative plausibility} we
    show how to map
    PPDs on $W$ to plausibility measures on $W$ in a way that preserves the
    truth-values of conditionals.
\end{itemize}

Plausibility structures are motivated by much the same concerns as two
other recent symbolic generalizations of probability by Darwiche
\citeyear{DarwicheThesis} and Weydert \citeyear{Weydert94}.  Their approaches
have a great deal more structure though.  They start with a domain $D$
and several algebraic operations that have properties similar to the
usual arithmetic operations (\eg addition and multiplication) over
$[0,1]$. The result is an algebraic structure over the domain $D$ that
satisfies various properties.  Their structures are also general
enough to capture all of the examples above except preferential
orderings.  These orderings cannot be captured precisely because of
the additional structure.  Moreover, as we shall see, by starting with
very little structure and adding just what we need, we can sometimes
bring to light issues that may be obscured in richer frameworks.  We
refer the interested reader to \cite{FrH7} for a more detailed
comparison to \cite{DarwicheThesis,Weydert94}.

Given the simplicity and generality of plausibility measures, we were
not surprised to discover that Weber \citeyear{Weber91} recently defined a
notion of {\em uncertainty measures}, which is a slight generalization
of plausibility measure
(in that domains more general than algebras of
sets are allowed), and that Greco \citeyear{Greco} defined a notion of
$L$-fuzzy measures which is somewhat more restricted than plausibility
measures in that the range $D$ is a complete lattice. We expect that
others have used variants of this notion as well, although we have not
found any further references in the literature.  To the best of our
knowledge, we are the first to carry out a systematic investigation of
the connection between plausibility measures and default reasoning.

\section{Approaches to Default Reasoning: A Review}
\label{default}
\label{DEFAULT}

Defaults are statements of the form ``if $\phi$ then
typically/likely/by default $\psi$'', which we denote $\phi \Cond
\psi$.  For example, the default ``birds typically fly'' is represented
$\Bird \Cond \Fly$. Formally, we assume that there is a ``base''
language $\L$, defined over a set $\Phi$ of propositions,
that includes the usual propositional connectives,
$\land,\lor,\neg,\rimp$ and has
a consequence relation $\vdashL$. The language of defaults $\LDef$
contains statements of the form $\phi \Cond \psi$, where $\phi, \psi
\in \L$.

There has been a great deal of discussion in the
literature as to what the appropriate semantics of defaults should be,
and what new defaults should be entailed by a knowledge base of
defaults.  For the most part, we do not get into these issues here.
While there has been little consensus on what the ``right'' semantics
for defaults should be, there has been some consensus on a reasonable
``core'' of inference rules for default reasoning.  This core, known
as the KLM properties \cite{KLM}, consists of
the following axiom and rules of inference.
\begin{description}\denselist
 \item[{\rm LLE}]
If $\vdashL \phi \dimp \phi'$, then from
 $\phi\Cond\psi$ infer
    $\phi'\Cond\psi$ (left logical equivalence)
 \item[{\rm RW}]
If $\vdashL \psi \rimp \psi'$, then from
 $\phi\Cond\psi$ infer
    $\phi\Cond\psi'$ (right weakening)
 \item[{\rm REF}] $\phi\Cond\phi$ (reflexivity)
 \item[{\rm AND}] From $\phi\Cond\psi_1$ and $\phi\Cond\psi_2$ infer
    $\phi\Cond \psi_1 \land \psi_2$
 \item[{\rm OR}] From $\phi_1\Cond\psi$ and $\phi_2\Cond\psi$ infer
    $\phi_1\lor\phi_2\Cond \psi$
 \item[{\rm CM}] From $\phi\Cond\psi_1$ and $\phi\Cond\psi_2$ infer
    $\phi\land \psi_2 \Cond \psi_1$ (cautious monotonicity)
\end{description}
LLE states that the syntactic form of
    the antecedent is irrelevant. Thus, if $\phi_1$ and $\phi_2$ are
    equivalent, we can deduce $\phi_2\Cond\psi$ from
    $\phi_1\Cond\psi$.  RW describes a
    similar property of the consequent: If $\psi$ (logically) entails
    $\psi'$, then we can deduce $\phi\Cond\psi'$ from $\phi\Cond\psi$.
    This allows us to can combine default and logical reasoning.
    REF states that $\phi$ is always a default conclusion
    of $\phi$. AND states that we can combine two default conclusions:
    If we can conclude by default both $\psi_1$ and $\psi_2$ from
    $\phi$, then we can also conclude  $\psi_1\land\psi_2$ from $\phi$.
    OR states that we are allowed to reason by cases: If the
    same default conclusion follows from each of two antecedents, then
it also follows from their disjunction. CM
    states that if $\psi_1$ and $\psi_2$ are two default
    conclusions of $\phi$, then discovering that $\psi_2$ holds
   when $\phi$
    holds (as would be
    expected, given the default) should not cause us to retract the
    default conclusion $\psi_1$.

This system of rules is called \SysP\ in \cite{KLM}. The notation
$\Delta \vdashp \phi\Cond\psi$ denotes that $\phi\Cond\psi$ can be deduced
    from $\Delta$ using these inference rules.

There are a number of well-known semantics for defaults that are
characterized by these rules.  We sketch a few of them here, referring
the reader to the original references for further details and
motivation.
All of these semantics involve structures of the form $(W,X,\pi)$,
    where $W$ is a set of possible worlds, $\pi(w)$ is a truth assignment
consistent with $\vdashL$ to formulas in $\L$,
and $X$ is some ``measure'' on $W$ such as
    a preference ordering, a $\kappa$-ranking,
    or a possibility measure.
    We define a little notation that will
    simplify the
    discussion below. Given a structure $M = (W ,X,\pi)$ and a formula
$\phi \in \L$, we take
    $\intension{\phi}_M \subseteq W$ to be the set of
worlds
    satisfying $\phi$, \ie $\intension{\phi}_M = \{ w \in W :
    \pi(w)(\phi) =$ {\bf true}$\}$. We omit the subscript $M$ when it
plays no role or
is clear from the context.

The first semantic proposal was provided by Kraus, Lehmann and Magidor
    \citeyear{KLM}, using ideas that go back to
    \cite{Hansson69,Lewis73,Shoham87}.
Recall that a preference ordering on $W$ is strict partial order (\ie
an
    irreflexive and transitive relation) $\prec$ over
    $W$.
A {\em preferential structure\/} is a tuple $(W,\prec,\pi)$,
    where $\prec$ is a strict partial order on $W$.%
\footnote{
We note that the formal definition of preferential
    structures in \cite{KLM,LehmannMagidor} is slightly more complex.
    Kraus, Lehmann, and Magidor distinguish between {\em states\/} and
    {\em worlds}. In their definition, a preferential structure is an
    ordering over states together with a labeling function that maps
    states to worlds. They take worlds to be truth assignments to
    primitive propositions. Our worlds thus
    correspond to states in
    their terminology, since we allow two worlds $w \neq w'$ such that
    $\pi(w) = \pi(w')$.
Despite these minor differences, all the results that we prove
for our version of preferential structures hold (with almost no
change in proof) for theirs.}
The intuition \cite{Shoham87} is that a preferential structure
     satisfies a conditional $\phi\Cond\psi$ if all the most
    preferred worlds (\ie the minimal worlds according to $\prec$) in
    $\intension{\phi}$ satisfy $\psi$.  However, there may be
no minimal worlds in $\intension{\phi}$. This can happen
    if $\intension{\phi}$ contains an infinite descending
    sequence $\ldots \prec w_2 \prec w_1$. What do we
    do in these structures?  There are
    a number of options: the first is to assume that,
for each formula $\phi$,
there are minimal worlds in
    $\intension{\phi}$
whenever $\intension{\phi}$ is not empty;
this is the
    assumption actually made in \cite{KLM}, where it is called the
    {\em smoothness\/} assumption.
A yet more general definition---one that works even if $\prec$ is
not smooth---is given in
    \cite{BossuSiegel85,Boutilier94AIJ1,Lewis73}.
    Roughly speaking, $\phi \Cond \psi$ is true if, from a certain
    point on, whenever $\phi$ is true, so is $\psi$.
More formally,
\begin{quote}
    $(W,\prec,\pi)$ satisfies  $\phi\Cond\psi$ if,
    for every
world $w_1 \in \intension{\phi}$,
    there is a world $w_2$ such that (a) $w_2 \preceq w_1$
(\ie either $w_2 \prec w_1$ or  $w_2 = w_1$ )
    (b) $w_2 \in \intension{\phi\land\psi}$, and (c) for all
    worlds $w_3 \prec w_2$, we have
     $w_3 \in \intension{\phi \rimp \psi}$ (so any world more
     preferred than $w_2$ that satisfies $\phi$ also satisfies $\psi$).
\end{quote}
It is easy to verify that this definition is equivalent to the
earlier one if $\prec$ is smooth.
A knowledge-base $\Delta$ {\em preferentially entails\/}
    $\phi\Cond\psi$, denoted $\Delta \satp\phi\Cond\psi$, if every
    preferential structure that satisfies (all the defaults in) $\Delta$
    also satisfies $\phi\Cond\psi$.

Lehmann and Magidor show that preferential entailment is characterized
by \SysP.
\thm {\rm \cite{LehmannMagidor,Boutilier94AIJ1}}
\label{thm:pref-complete}
$\Delta \satp \phi\Cond\psi$ if and only if $\Delta \vdashp
    \phi\Cond\psi$.
\ethm
Thus, reasoning with preferential structures corresponds in a precise
    sense to reasoning with the core properties of default reasoning.

As we mentioned earlier, we usually want to add additional inferences
to those sanctioned by the core. Lehmann and Magidor
\citeyear{LehmannMagidor} hoped to do so by
restricting to a
special class of preferential structures. A preferential structure
$(\Omega,\prec, \pi)$ is {\em rational\/} if $\prec$ is a {\em
modular\/} order, so that for all worlds $u,v, w \in \Omega$, if $w
\prec v$, then either $u \prec v$ or $w \prec u$.  It is not hard to
show that modularity implies that possible worlds are clustered into
equivalence classes, each class consisting of worlds that are
incomparable to one another, with these classes being totally ordered.
Thus, rational structures form a ``well-behaved'' subset of
preferential structures.  Unfortunately, Lehmann and Magidor showed
that restricting to rational structures gives no additional properties
(at least, as far as the limited language of defaults is concerned).
We say that
a knowledge base $\Delta$ {\em rationally entails\/}
    $\phi\Cond\psi$, denoted $\Delta \satr\phi\Cond\psi$, if every
    rational structure that satisfies $\Delta$
    also satisfies $\phi\Cond\psi$.%
\footnote{Rational entailment should not be confused with the notion
of {\em rational closure}, also defined by Lehmann and Magidor \citeyear{LehmannMagidor}.}
\thm{\rm \cite{LehmannMagidor}}
\label{thm:rational-complete}
$\Delta \satr \phi\Cond\psi$ if and only if $\Delta \vdashp
    \phi\Cond\psi$.
\ethm
Thus, we do not gain any new patterns of default inference when we
    restrict our attention to rational structures.

This is perhaps somewhat surprising, since it is is known that rational
structures do satisfy the following additional property, known as {\em
rational monotonicity} \cite{KLM,LehmannMagidor}:
\begin{itemize}
\item[{\rm RM}] If $\phi \Cond \psi_1$ and $\phi 
\nCond \neg \psi_2$
then
$\phi \land \psi_2 \Cond \psi_1$.
\end{itemize}
Note that RM is almost the same as CM, except that $\phi \Cond \psi_2$
is replaced by the weaker 
$\phi \nCond \neg \psi_2$.

How can the
existence of this additional property be consistent with the fact both
that rational and preferential structures are characterized by \SysP?
The key point is that although RM is an additional property satisfied by
rational structures, it is not one that is expressible in the language
of defaults (because we do not allow negated defaults).  As we shall see
in Section~\ref{conditionallogic}, once we move to a richer language,
rational
structures are distinguishable form arbitrary preferential structures.

Pearl \citeyear{Pearl90} considers a semantics for defaults grounded in
probability, using an approach due to Adams \citeyear{Adams:75}.  In this
approach, a default $\phi\Cond\psi$ denotes that
$\Pr(\psi | \phi)$
is extremely high, \ie almost $1$.  Roughly speaking, a collection
$\Delta$ of defaults implies a default $\phi \Cond \psi$ if we can
ensure that $\Pr(\phi|\psi)$ is arbitrarily close to 1, by taking the
probabilities of the defaults in $\Delta$ to be sufficiently high.

The formal definition needs a bit of machinery.%
\footnote{We adopt the presentation used in \cite{GMPfull}.}
Recall that a PPD on $W$ is a sequence $ \{\Pr_i : i \ge 0\}$ of
probability measures over $W$.  A {\em PPD structure\/} is a tuple
$(W,\{ \Pr_i : i \ge 0 \},\pi)$, where $\{\Pr_i\}$ is PPD on $W$.
Intuitively, it satisfies a conditional $\phi\Cond\psi$ if the
conditional probability $\psi$ given $\phi$ goes to $1$ in the limit.
Formally, $\phi\Cond\psi$ is satisfied if $\lim_{i \rightarrow
 \infty}\Pr_i(\intension{\psi}|\intension{\phi}) = 1$
(where $\Pr_i(\intension{\psi}|\intension{\phi})$ is taken to be 1 if
$\Pr_i(\intension{\phi}) = 0$).  $\Delta$ {\em $\epsilon$-entails\/}
$\phi\Cond\psi$, denoted $\Delta\sate\phi\Cond\psi$, if every PPD
structure that satisfies all the defaults in $\Delta$ also satisfies
$\phi\Cond\psi$.  Surprisingly, Geffner shows that
$\epsilon$-entailment is equivalent to preferential entailment.

\thm {\rm \cite{Geffner92}}
\label{thm:epsilon-complete}
$\Delta\sate\phi\Cond\psi$ if and only if $\Delta\vdashp\phi\Cond\phi$.%
\footnote{Geffner's result is stated in terms of the original
formulation of $\epsilon$-entailment, as described in
\cite{Pearl90}. However, results of \cite{GMPfull} show that the
formulation we describe here is equivalent to the original one.}
\ethm

Possibility measures and ordinal rankings provide two more semantics
for defaults.  A {\em possibility structure\/} is a tuple $\PS =
(\Omega,\Poss,\pi)$ such that $\Poss$ is a possibility measure on
$\Omega$.  We say $\PS \satPoss \phi \Cond \psi$ if either
$\Poss(\intension{\phi}) = 0$ or $\Poss(\intension{\phi
\land \psi}) > \Poss(\intension{\phi \land \neg \psi})$.
Intuitively, $\phi \Cond \psi$ holds vacuously if $\phi$ is
impossible; otherwise, it holds if $\phi \land \psi$ is more
``possible'' than $\phi \land \neg \psi$.  For example, $\Bird \Cond
\Fly$ is satisfied when $\Bird \land \Fly$ is more possible than
$\Bird \land \neg \Fly$.  Similarly, an {\em ordinal ranking
structure\/} is a tuple $R = (\Omega,\kappa,\pi)$ if $\kappa$ is an
ordinal ranking on $\Omega$.  We say that $R \satkappa \phi \Cond
\psi$ if either $\kappa(\intension{\phi}) = \infty$ or
$\kappa(\intension{\phi\land \psi}) < \kappa(\intension{\phi \land
\neg \psi})$.  We say that $\Delta$ {\em possibilistically entails\/}
$\phi \Cond \psi$, denoted $\Delta \sat_{\sPoss} \phi \Cond \psi$
(\respc $\Delta$ {\em $\kappa$-entails\/} $\phi \Cond \psi$, denoted
$\Delta \satkappa \phi \Cond \psi$) if all possibility structures
(\respc all ordinal ranking structures) that satisfy $\Delta$ also
satisfy $\phi \Cond \psi$.

These two approaches are again characterized by the KLM properties.
\thm {\rm \cite{Geffner92,DuboisPrade:Defaults91}}
The following are equivalent:
\begin{itemize}
\item[(a)]
$\Delta \satPoss \phi \Cond \psi$
\item[(b)] $\Delta \satkappa \phi
\Cond \psi$
\item[(c)] $\Delta\vdashp\phi\Cond\psi$.
\end{itemize}
\ethm

Why do we always seem to end up with the KLM properties?  As we are
about to show, thinking in terms of plausibility measures
provides the key to understanding this issue.

\section{Default Reasoning Using Plausibility}
\label{qualitative plausibility}
\label{QUALITATIVE PLAUSIBILITY}

We can give semantics to defaults using plausibility measures much as
we did using possibility measures.  A {\em plausibility structure\/}
(for $\L$)
is a tuple $\PL = (\Omega, \F, \Pl, \pi)$, where $(\Omega,\F, \Pl) $ is a
plausibility space and $\pi$ maps each possible world to a truth
assignment
to the formulas in $\L$ that is consistent with $\vdashL$ in the
obvious sense.
Since we will be interested in events that correspond to formulas, we
require that $\intension{\phi} \in \F$ for all formula $\phi \in
\L$.
For ease of exposition,
when describing a plausibility structure for $\L$, we assume that
$\F = \{ \intension{\phi} : \phi \in \L \}$.
Just as with plausibility spaces, we typically omit the
algebra $\F$ from the description of a plausibility structure.
We define
    $\PL \satPL \phi\Cond\psi$ if either $\Pl(\intension{\phi}) =
\bottom$ or
 $\Pl(\intension{\phi\land\psi}) >
    \Pl(\intension{\phi\land\neg\psi})$.

Notice that if $\Pl$ is a probability function $\Pr$, then $\phi \Cond
\psi$ holds exactly if either $\Pr(\intension{\phi}) = 0$ or
$\Pr(\intension{\psi}|\intension{\phi}) > 1/2$.  How does this
semantics for defaults compare to others that have been given in the
literature?  It is immediate from the definitions that the semantics
we give to defaults in possibility structures is the same as that
given to them if we view these possibility structures as plausibility
structures (using the obvious mapping described in
Section~\ref{default}), and similarly for ordinal ranking structures.
What about preferential structures and PPD structures?  Can we map
them into plausibility structures while still preserving the semantics
of defaults?  As we now show, we can.

In fact, \Lref{lem:embedding} shows that there is a general procedure
for mapping any approach that satisfies the KLM postulates to
plausibility measures. Before describing this general construction, we
briefly sketch its instantiation in the case of PPDs and preferential
structures.

We start with PPDs. Let $PP = (W, \{ \Pr_i \}, \pi)$ be a
PPD structure. Let $\Pl_{PP}$ be a plausibility measure on $W$ such that
\beqn
\Pl_{PP}(A) \le \Pl_{PP}(B)\mbox{ if and only if }\lim_{i \rightarrow
    \infty}{\Pr}_i(B|A \union B) =  1.
\label{eq:PPD-embed}
\eeqn
It is easy to check that such a plausibility measure exists and that
$(W, \Pl_{PP}, \pi)$ satisfies the same defaults as $PP$.
We note that this construction, as well as others in the remainder of
the paper, specifies only the relative order of plausibilities of
events, and does not describe the domain of plausibility values. It is
easy to check that as long as the ordering constraints are consistent
with reflexivity, transitivity, and A1, we can always construct a
matching plausibility domain.%
\footnote{For example,
we can take the domain of the plausibility measure to consist of
sets of logically equivalent formulas, partially ordered so as to
satisfy the constraints.}
{From} here on, we treat such ordering
constraints as though they define a plausibility measure.

We stress that this embedding, which is sufficient for the purpose of
this work, is not the only one possible. To see this, suppose that $A$
and $B$ are disjoint sets such that $\Pr_i(A) = \Pr_i(B)$ for all $i$.
One might argue that the plausibility of $A$ and $B$ should be equal.
Yet our definition would make $\Pl(A)$ and $\Pl(B)$ incomparable since
$\Pr_i(B | A \union B) = 0.5$ for all $i$.

The construction for mapping preferential structures into plausibility
structures is slightly more
complex.
Suppose we are given a preferential structure $(W,\prec,\pi)$.
Let $D_0$ be the domain of
    plausibility values consisting of one element $d_w$ for every
    element $w \in W$.  We use $\prec$ to determine the order of these
    elements: $d_v < d_w$ if $w \prec v$. (Recall that $w \prec w'$
    denotes that $w$ is preferred to $w'$.)  We then take $D$ to be
    the smallest set containing $D_0$ closed under least upper bounds
    (so that every set of elements in $D$ has a least upper bound in
    $D$).
It is not hard to show that $D$ is well-defined (\ie there is a
unique, up to renaming, smallest set) and that taking $\Pl_\prec(A)$ to
be the least upper bound of $\{ d_w : w \in A \}$ gives us the
following property:
\beqn
\parbox{4.2in}{
$\Pl_\prec(A) \le \Pl_\prec(B)$ if and only if for all $w \in A - B$,
    there is a world $w' \in B$ such that $w' \prec w$ and there is no
    $w'' \in A - B$ such that $w'' \prec w'$.}
\label{eq:prec-embed}
\eeqn
Again, it is easy to check that $(W,\Pl_\prec,\pi)$ satisfies the same
defaults as $(W,\prec,\pi)$.

We now present our general construction.

\lem\label{lem:embedding}
Let $W$ be a set of possible worlds and let $\pi$ be a function that maps each
world in $W$ to a truth assignment to $\L$.
Let $T \subseteq \LDef$ be a set of defaults that is closed under the
rules of \SysP\ that satisfies the following condition:
\begin{quote}
{\rm ($*$)} if $\phi \Cond \psi \in T$, $\intension{\phi} = \intension{\phi'}$, and
$\intension{\psi} = \intension{\psi'}$,  then $\phi' \Cond \psi' \in T$, for
all formulas $\phi, \phi', \psi, \psi' \in \L$.
\end{quote}
There is a plausibility structure $\PL_T = (W,\Pl_T, \pi)$ such that
$\Pl_T(\intension{\phi}) \le \Pl_T(\intension{\psi})$ if and
only if $\phi\lor\psi \Cond \psi \in T$.
Moreover,
$\PL_T \sat \phi \Cond \psi$ if and only if $\phi \Cond \psi \in T$.
\elem
\prf
See Appendix~\ref{prf:qualplaus}.
\eprf

\thm\label{thm:plausembedding}
\begin{itemize}\denselist
 \item[(a)] Let $PP = \{ \Pr_i \}$ be a PPD on $W$.  There is a
plausibility measure $\Pl_{PP}$ on $W$ such that $(W,\{\Pr_i\}, \pi)
\sat_\epsilon \phi\Cond\psi$ if and only if $(W,\Pl_{PP},\pi)
\sat_{\sPL} \phi\Cond\psi$.
 \item[(b)] Let $\prec$ be a preference ordering on $W$.  There is a
plausibility measure $\Pl_\prec$ on $W$ such that $(W,\prec,\pi)
\sat_p \phi\Cond\psi$ if and only if $(W,\Pl_\prec,\pi) \sat_{\sPL}
\phi\Cond\psi$.
\end{itemize}
\ethm

\prf
We start with part (a). Set $T_{PP} = \{ \phi \Cond\psi :
(W,PP,\pi) \sat_\epsilon \phi \Cond \psi \}$. \Tref{thm:epsilon-complete}
implies that $T_{PP}$ is closed under the KLM rules. Moreover, since
$T$ is constructed from a PPD over $W$ using $\pi$ it
satisfies the requirement of \Lref{lem:embedding}. It follows that we
can construct a plausibility structure that satisfies the requirements
of the theorem.
It is also easy to verify that this construction
agrees with the construction described earlier, in that it satisfies
constraint
(\ref{eq:PPD-embed}). To see this,
let $A, B \in \F$. Then, according to our
assumptions, there are formulas $\phi$ and $\psi$ such that $A =
\intension{\phi}$ and $B = \intension{\psi}$. By definition,
$\Pl_{PP}(A) \le \Pl_{PP}(B)$ if and
only if $(W,PP,\pi) \sat_\epsilon (\phi\lor\psi) \Cond \psi$. From
definition of $\sat_\epsilon$ we immediately get (\ref{eq:PPD-embed}).

The proof of part (b) is identical, using \Tref{thm:pref-complete}.
Again, an analogous argument easily shows that $\Pl_\prec$ satisfies
(\ref{eq:prec-embed}).
\eprf

Thus, each of the semantic approaches to default reasoning that were
considered in Section~\ref{default} can be mapped into plausibility
structures in a way that preserves the semantics of defaults.
We remark that these mapping are not unique.  For example, Freund
\citeyear{Freund96} gives an alternative mapping from preferential
structures to plausibility measures.

Other semantics for defaults can also be mapped into
plausibility measures using the general technique of
\Lref{lem:embedding}.
In most cases, we can also establish a direct
relationship between these semantics and plausibility measures.
For
example,
the {\em coherent filters\/} approach of
\cite{BenDavidBenEliyahu,Schlechta95} can be mapped to
plausibility, as shown by
Schlechta \citeyear{Schlechta:Plaus},
and
Weydert's {\em full ranking measures\/}
\citeyear{Weydert94b} are easily seen to be
a special case of plausibility measures.

\section{Default Entailment in Plausibility Structures}
\label{axioms}
\label{AXIOMS}
In this section we characterize default entailment in plausibility
structures.  To do so, it is useful to have a somewhat more general
definition of entailment in plausibility
structures.
\dfn
If $\PLclass$ is a class of plausibility structures, then a knowledge base
$\Delta$ entails $\phi\Cond\psi$ with respect to $\PLclass$, denoted
$\Delta\sat_\PLclass \phi\Cond\psi$, if every plausibility structure $PL \in
\PLclass$ that satisfies all the defaults in $\Delta$ also satisfies
$\phi\Cond\psi$.
\edfn
The classes of structures we are interested in include $\PLclass^{PL}$, the
class of all plausibility structures, and $\PLclass^{Poss}$,
$\PLclass^\kappa$, $\PLclass^p$, $\PLclass^r$, and $\PLclass^{\epsilon}$, the classes that arise
{from } mapping
possibility structures, ordinal ranking structures,
preferential structures, rational structures, and PPDs, respectively,
into plausibility
structures.  (In the case of possibility structures and ordinal ranking
structures, the mapping is the obvious one discussed in
Section~\ref{plausibility}; in the case of
preferential, rational and  PPD structures, the mapping is the one
described in Theorem~\ref{thm:plausembedding}.)  Recall that all these
mappings
preserve the semantics of defaults.

It is easy to check that our semantics for defaults does {\em not\/}
guarantee that the axioms of \SysP\ hold in all structures in
$\PLclass^{PL}$.  In particular, they do not hold in probability
structures.
For a counterexample, consider a plausibility structure $PL =
(W,\Pl,\pi)$, where $\Pl$ is actually a probability measure $\Pr$ such
that
$\Pr(\intension{q\land r}) = 0.2$ and
$\Pr(\intension{q \land \neg r}) = \Pr(\intension{\neg q
\land r}) = 0.4$.  Thus, $\Pr(\intension{q}) = .6$ and
$\Pr(\intension{q}|\intension{r}) = 1/3$.   Recall that
if $\Pr(\intension{\phi}) > 0$, then $\PL \satPL \phi \Cond \psi$
if and only if $\Pr(\intension{\psi}|\intension{\phi}) > .5$.
Thus, $\PL \satPL (\true \Cond q) \land (\true \Cond r)$,
but $\PL {\not\sat}_{PL} \true \Cond (q\land r)$ and
$\PL {\not\sat}_{PL} r \Cond q$.
This gives us a violation of both AND and CM.  We can similarly
construct a counterexample to OR.  On the other hand, as the following
result shows, plausibility structures do satisfy the other three axioms
of \SysP.  Let \SysP$'$ be the system consisting of LLE, RW, and
REF.

\thm\label{thm:SysP'}
If $\Delta \vdashpp \phi\Cond\psi$, then $\Delta \sat_{\PLclass^{PL}} \phi\Cond\psi$.
\ethm
\prf
See Appendix~\ref{prf:axioms}.
\eprf

What extra conditions do we have to place on plausibility structures
to ensure that AND, OR, and CM are satisfied?  We focus first on the
AND rule.  We want an axiom that cuts out probability functions, but
leaves more qualitative notions.  Working at a semantic level, taking
$\intension{\phi} = A$, $\intension{\psi_1} = B_1$, and
$\intension{\psi_2} = B_2$, and using $\overline{X}$ to denote the
complement of $X$, the AND rule translates to
\begin{itemize}
\item[\bf A2$'$.] For all sets $A$, $B_1$, and $B_2$, if
$\Pl(A \inter B_1) > \Pl(A \inter \overline{B_1})$
and $\Pl(A \inter B_2) > \Pl(A \inter \overline{B_2})$, then
$\Pl(A \inter B_1 \inter B_2) > \Pl(A \inter \overline{(B_1 \inter
B_2)})$.
\end{itemize}
It turns out that in the presence of A1, the following somewhat
simpler axiom is equivalent to A2$'$:
\begin{itemize}\denselist
 \item [\bf A2.] If $A$, $B$, and $C$ are pairwise disjoint sets,
 $\Pl(A \union B) > \Pl(C)$, and $\Pl(A \union C) > \Pl(B)$, then
$\Pl(A) > \Pl(B \union C)$.
\end{itemize}

\pro\label{pro:A2} A plausibility measure satisfies A2 if and only if it satisfies A2$'$.
\epro
\prf
See Appendix~\ref{prf:axioms}.
\eprf

A2 can be viewed as a generalization
of a natural requirement of qualitative
plausibility:
if $A$, $B$, and
$C$ are pairwise disjoint,
$\Pl(A) > \Pl(B)$, and $\Pl(A) > \Pl(C)$,
then $\Pl(A) > \Pl(B \union C)$.
Moreover, since A2 is equivalent to A2$'$, and A2$'$ is a direct
translation of the AND rule into conditions on plausibility
measures, any plausibility structure whose plausibility measure
satisfies A2 also satisfies the AND rule.
Somewhat surprisingly, a plausibility measure $\Pl$ that satisfies A2
also satisfies CM. Moreover, $\Pl$ satisfies the non-vacuous case of the OR
    rule. That is, if $\Pl(\intension{\phi_1}) > \bot$, then from
    $\phi_1 \Cond \psi$ and $\phi_2 \Cond\psi$ we can conclude
    $(\phi_1\lor\phi_2) \Cond\psi$.%
\footnote{We remark that if we dropped requirement A1, then
we can define properties of plausibilities measures that
correspond precisely to CM and OR.  The point is that in
the presence of A1,
A2---which essentially corresponds to AND---implies CM
and the non-vacuous case of OR.
Despite appearances, A1 does {\em not\/} correspond to RW.
Semantically, RW says that if $A$ and $B$ are disjoint sets
such that $\Pl(A) > \Pl(B)$, and $A \subseteq A'$, $B' \subseteq
B$, and $A'$ and $B'$ are disjoint, then $\Pl(A') > \Pl(B')$.
While this follows from A1, it is much weaker than A1.}
To handle the vacuous case of OR we need an additional axiom:
\begin{itemize}\denselist
 \item [\bf A3.] If $\Pl(A) = \Pl(B) = \bot$, then $\Pl(A \union B) = \bot$.
\end{itemize}
Thus, A2 and A3 capture the essence of the KLM properties.
To make this precise, define a plausibility space $(\Omega,\Pl)$ to be
{\em qualitative\/} if it satisfies A2 and A3 in addition to A1.
We say $\PL = (\Omega,\Pl,\pi)$ is a {\em qualitative plausibility
structure\/} if $(\Omega,\Pl)$ is a qualitative plausibility
space.
Let $\PLclass^{QPL}$ consist of
all qualitative plausibility structures.

\thm\label{QPL}
$\PLclass \subseteq \PLclass^{QPL}$ if and only if  for
all $\Delta$, $\phi$, and $\psi$,
if $\Delta\vdashp
    \phi\Cond\psi$ then $\Delta\sat_\PLclass \phi\Cond\psi$.
\ethm
\prf
See Appendix~\ref{prf:axioms}.
\eprf

Thus, the KLM axioms are sound for qualitative plausibility
structures.
We remark that Theorem~\ref{QPL}
provides not only a
sufficient but a necessary
condition for a set of plausibility structures to satisfy the KLM
properties:
If the KLM axioms are sound with respect to $\PLclass$,
then
all $\PL \in \PLclass$ must be qualitative.

This, of course, leads to the
question of which plausibility structures are qualitative.
All the ones we have been focusing on are.

\thm\label{everythinginQPL}
Each of
$\PLclass^{Poss}$, $\PLclass^\kappa$, $\PLclass^\epsilon$, $\PLclass^p$, and $\PLclass^r$ is a
subset of  $\PLclass^{QPL}$. \ethm
\prf
See Appendix~\ref{prf:axioms}.
\eprf

It follows from Theorems~\ref{QPL} and~\ref{everythinginQPL} that the
KLM properties hold in all the approaches to default reasoning
considered in Section~\ref{default}.  While this fact was already
known, this result gives us a deeper understanding as to {\em why\/}
the KLM properties should hold.
In a precise sense, it is because A2 and A3 hold for all these
approaches.

We now consider completeness.  To get soundness,
we have to ensure that $\PLclass$ does not contain too many structures,
in particular, no structures that are not qualitative.
To get completeness, we have to ensure that $\PLclass$ contains ``enough''
structures.  In particular, if $\Delta \not\vdashp \phi \Cond
\psi$, we want to ensure that there is a plausibility structure $PL \in
\PLclass$ such that $PL \satPL \Delta$ and
$PL {\not\sat}_{PL} \phi \Cond \psi$.  The following
weak condition on $\PLclass$ does this.
\dfn\label{dfn:richness}
We say that $\PLclass$ is {\em rich\/} if for every collection $\phi_1,
\ldots, \phi_n$, $n > 1$, of mutually exclusive formulas,
there is a plausibility structure $PL = (\Omega,\Pl,\pi) \in \PLclass$ such
that:
$$
\Pl(\intension{\phi_1}) > \Pl(\intension{\phi_2}) >
\cdots > \Pl(\intension{\phi_n}) = \bot. \ \bbox$$
\end{definition}
\commentout{
We say that $\PLclass$ is {\em rich\/} if for every collection of formulas
$\{ \phi_{i,j} : 1 \le i \le n, 1 \le j \}$ such that $\vdash
    \phi_{i,j+1} \rimp \phi_{i,j}$ for all $1 \le i < n$ and $1 \le j$, $\vdash
    \phi_{n,j} \rimp \phi_{n,j+1}$ for all $1 \le j$ and $\vdashL \neg
(\phi_{i,j} \land \phi_{l,j})$ for all $i \neq l$ and $1 \le j$, there
is a plausibility structure $PL = (\Omega,\Pl,\pi) \in \PLclass$ such
that:
$$
\Pl(\intension{\phi_{1,j}}) > \Pl( \intension{\phi_{2,j}}) >
\cdots > \Pl(\intension{\phi_{n,j}}) = \bot. \mbox{ for all $1 \le j$}
$$
\end{definition}

This definition might seem somewhat complex. However, when the language
$\L$ has finite number of propositions, richness has a simpler formulation:
\pro
Let $\L$ be a finite language. Then $\PLclass$ is rich if and only
if for every collection $\phi_1,
\ldots, \phi_n$, $n > 1$, of mutually exclusive formulas,
there is a plausibility structure $PL = (\Omega,\Pl,\pi) \in \PLclass$ such
that:
$$
\Pl(\intension{\phi_1}) > \Pl(\intension{\phi_2}) >
\cdots > \Pl(\intension{\phi_n}) = \bot.
$$
\epro
}

The richness requirement is quite mild.  It says that we do not
have {\em a priori\/} constraints on the relative plausibilities of a
collection of disjoint sets.  
Theorem~\ref{allOK} shows that
every collection of
plausibility measures that we have considered thus far can be easily shown
to satisfy this richness condition.
More importantly, Theorem~\ref{thm:main} shows that richness is a
necessary and sufficient condition to 
ensure that the KLM properties are complete.

\thm\label{allOK} Each of $\PLclass^{Poss}$, $\PLclass^\kappa$, $\PLclass^p$, $\PLclass^r$, $\PLclass^{\epsilon}$, and
$\PLclass^{QPL}$ is rich. \ethm
\prf
Let $\phi_1, \ldots, \phi_n$, $n > 1$, be 
mutually
exclusive formulas and let $W = \{ w_1, \ldots, w_{n-1}
\}$. Since $\phi_1, \ldots, \phi_n$ are mutually exclusive, we can
construct a mapping $\pi$ that maps each world in $W$ to a truth
assignment such that $\intension{\phi_i} = w_i$ for all $1 \le i < n$, and
$\intension{\phi_n} = \emptyset$. Recall that we need to find a
plausibility measure $\Pl$ 
such that 
$\Pl(\intension{\phi_1}) >
\Pl(\intension{\phi_2}) > \cdots > \Pl(\intension{\phi_n}) = \bot$. 
It is easy to find a plausibility measure $\Pl$ satisfying this property
such that 
$(W,\Pl, \pi)$ is in $\PLclass^{Poss}$, $\PLclass^\kappa$,
$\PLclass^p$, $\PLclass^r$, $\PLclass^{\epsilon}$, or
$\PLclass^{QPL}$.  For example, to get a structure in
$\PLclass^{Poss}$, we define $\Poss(w_i) = 1 - \frac{i}{n}$. To get a
structure in $\PLclass^{p}$, we define $\Pl$ to correspond to the
preference ordering $w_1 \prec w_2 \prec \ldots \prec w_{n-1}$.
\eprf

\thm
\label{thm:main}
A set $\PLclass$ of qualitative plausibility structures is rich if and only
if for all
finite $\Delta$ and defaults $\phi \Cond \psi$, we have that
$\Delta\sat_\PLclass\phi\Cond\psi$ implies $\Delta\vdashp\phi\Cond\psi$.
\ethm
\prf
See Appendix~\ref{prf:axioms}.
\eprf

Note that \Tref{thm:main} deals with what is usually considered to be
{\em weak\/} completeness. The {\em strong\/} notion of completeness would
require us to remove the restriction that $\Delta$ is finite from the
statement of the theorem.
It is possible to find
a stronger notion of richness that corresponds to strong
completeness, but the details are somewhat cumbersome, so we
do not provide them here.
Note that if $\sat_\PLclass$ is compact,
then weak completeness implies strong completeness.

Putting together Theorems~\ref{QPL}, \ref{everythinginQPL},
and~\ref{thm:main}, we get
\cor For $\PLclass \in \{
\PLclass^{Poss}, \PLclass^\kappa, \PLclass^p, \PLclass^r, \PLclass^{\epsilon},
\PLclass^{QPL}\}$, and
all $\Delta, \phi$, and $\psi$, we have $\Delta\vdashp\phi\Cond\psi$
    if and only if $\Delta\sat_\PLclass\phi\Cond\psi$. \ecor

Not only does this result gives us a straightforward and uniform proof
that the KLM properties characterize default reasoning in each of the
systems considered in Section~\ref{default}, it gives us a general
technique for proving completeness of the KLM properties for other
semantics as well.
All we have to do is to provide
a mapping
of the
    intended semantics into plausibility structures, which is usually
    straightforward, and then show that the resulting set of structures
is
    qualitative and rich.

Theorem~\ref{thm:main} also has important implications for attempts to
go beyond the KLM properties (as was the goal in introducing rational
structures).  It says that any semantics
for defaults that proceeds by considering a class $\PLclass$ of
qualitative structures
satisfying the richness constraint,
and defining $\Delta \sat_\PLclass \phi \Cond \psi$ to hold if
$\phi \Cond \psi$ is true in every structure in $\PLclass$ that satisfies
$\Delta$, cannot lead to new
properties for
    entailment.

Thus, to go beyond KLM, we
need to either consider interesting
non-rich classes of structures or to define a notion of entailment
from $\Delta$
that does not consider all the structures
of a given class.
We are not aware of any work that takes the first approach, although
it is possible to construct classes of structures
that are arguably interesting and violate the richness constraint.
One way is to impose independence constraints.  For example, suppose
the language includes primitive propositions $p$ and $q$, and
we consider all structures where $p$ is independent of $q$ in the
sense that if any of $\true \Cond q$, $p \Cond q$, and $\neg p \Cond
q$ holds, then the others also do.
This means
that discovering either $p$ or
$\neg p$ does not affect whether or not $q$ is believed.%
\footnote{We remark that if we define independence appropriately
in plausibility structures, this property does indeed hold; see
\cite{FrH7}.} Restricting to such structures clearly gives us extra
properties.  For example, from $\true \Cond q$ we can infer $p \Cond
q$, which certainly does not follow from the KLM properties.  Such
structures do not satisfy the richness constraint, since we cannot
have, for example, $\Pl(\intension{p \land q} ) > \Pl(\intension{p
\land \neg q}) > \Pl(\intension{\neg p \land \neg q}) >
\Pl(\intension{\neg p \land q})$.

Much of the recent work in default reasoning
\cite{BGHKnonmon,geffner:92,Goldszmidt92,GMPfull,LehmannMagidor,Pearl90B}
has taken the second approach.  Roughly speaking, this approach
can be viewed as taking the basic idea of preferential
semantics---placing a preference ordering on worlds---one step
further: We try to get from a knowledge base a set of preferred {\em
structures} (where the structures themselves put a preference ordering on
worlds)---for example, in \cite{GMPfull}, the PPD of maximum entropy
is considered---and carry out all reasoning in these preferred
structures.  We believe that plausibility measures will provide insight
into techniques for choosing such preferred structures.
For example, we might want to prefer structures where things are ``as
independent as possible''.  We believe that it should be possible to
capture this notion in a reasonable way 
using plausibility; we defer this to future work.
(See \cite{FrH7} for discussion on independence in the context of
plausibility.)

\section{Expressiveness of Qualitative Plausibility
Measures}\label{sec:epsilon}\label{SEC:EPSILON}

In the previous section we showed that all approaches to default
reasoning are instances of qualitative plausibility structures.
We now show that each of the classes considered in
    \Tref{everythinginQPL} is a strict subset of $\PLclass^{QPL}$.
This is clearly true in a trivial sense.  For example, if we consider a
qualitative plausibility measure whose range is $[1,2]$, it cannot
be either a possibility measure or a $\kappa$-ranking.
To get around this problem, we define two plausibility spaces
$(W,\Pl)$ and $(W,\Pl')$ (\resp, two plausibility structures $(W,
\Pl,\pi)$ and $(W,\Pl',\pi)$) to be
{\em order-equivalent\/}
if for $A, B \subseteq W$, we have $\Pl(A) \le \Pl(B)$ if and only
if
$\Pl'(A) \le \Pl'(B)$.

We claim that for each
of the classes of plausibility
structures
considered in \Tref{everythinginQPL}, there
is a qualitative plausibility
structure that is not order-equivalent to any
element of that class.
This is almost immediate in the case of
$\PLclass^{Poss}$ and $\PLclass^\kappa$.  Since both require
$\le$ to be a total order, a qualitative plausibility
structure $(W,\Pl,\pi)$ such that $\Pl$
does not place a total order on the plausibility of subsets cannot be
order-equivalent to an element of $\PLclass^{Poss}$ or $\PLclass^\kappa$.
We say a plausibility structures $(W,\Pl,\pi)$ is {\em totally
ordered\/} if $\Pl$ places a total order on subsets.
As the following proposition show, there are even totally-ordered
qualitative plausibility structures that are not order-equivalent
to any possibility structure or ordinal ranking structure.
\pro\label{pro:order-iso}
There is a totally-ordered qualitative plausibility structure that
is not order-equivalent to any structure in $\PLclass^{Poss}$,
$\PLclass^\kappa$,
$\PLclass^p$, or $\PLclass^{\epsilon}$.
\epro
\prf
Define a plausibility measure $\Pl$ on
$\{a,b,c\}$
such that $\Pl(\{a\}) = \Pl(\{b\}) = \Pl(\{c\}) = \Pl(\{b,c\}) =
\Pl(\{a,c\}) = 1/2$ and $\Pl(\{a,b\}) = \Pl(\{a,b,c\}) = 1$.
It is straightforward to check that $\Pl$ is qualitative and
totally ordered. Moreover, we have
$\Pl(\{c\}) < \Pl(\{a,b\})$, although neither
$\Pl(\{c\}) < \Pl(\{a\})$ nor $\Pl(\{c\}) < \Pl(\{b\})$ hold.
It is easy to see that there can be no possibility measure,
$\kappa$-ranking, preference ordering, or PPD on $\{a,b,c\}$ such
that the corresponding plausibility space is order-equivalent to
$(\{a,b,c\},\Pl)$. For example, if $\Poss$ is a possibility measure on
$\{a,b,c\}$ such that $\Poss(\{c\}) < \Poss(\{a,b\}$, then we must have
either $\Poss(\{c\}) < \Poss(\{a\}$ or $\Poss(\{c\}) < \Poss(\{b\})$.
A similar observation holds for $\kappa$-rankings.
This plausibility space also cannot be equivalent to one that arises from the
construction of \Lref{lem:embedding},
since the construction never gives disjoint sets the same plausibility.
Since
$\Pl(\{a\}) = \Pl(\{b\})$,
the result follows. \eprf

If all that we are interested in is default reasoning, then all that
matters is the relative plausibility of disjoint sets.  We say that
two plausibility spaces $(W,\Pl)$ and $(W,\Pl')$ (\resp two
plausibility structures $(W,\Pl,\pi)$ and $(W,\Pl',\pi)$) are
{\em default-equivalent\/}
if for all disjoint subsets $A$ and $B$ of $W$, we have
$\Pl(A) < \Pl(B)$ if and only if $\Pl'(A) < \Pl'(B)$.
Clearly, if
structures $(W,\Pl,\pi)$
and $(W,\Pl',\pi)$
are default-equivalent, then they
satisfy the same defaults.

We can strengthen Proposition~\ref{pro:order-iso} so that it applies to
default-equivalence in the case of possibility measures,
$\kappa$-rankings, and preferential orders.
\pro\label{pro:default-iso}
There is a totally-ordered qualitative plausibility structure that
is not default-equivalent to any structure in $\PLclass^{Poss}$,
$\PLclass^\kappa$, or $\PLclass^p$.
\epro
\prf
The plausibility space described in the proof of
\Pref{pro:order-iso}
also provides
a counterexample for default-equivalence
in the case of $\PLclass^{Poss}$, $\PLclass^{\kappa}$, and $\PLclass^p$.
\eprf

\commentout{We note that for finite algebras of events, we can similar
structures
in $\PLclass^{QPL}$ by structures in $\PLclass^p$ by adding additional worlds.
This is an straightforward conclusion from the construction of Kraus,
Lehmann and Magidor \citeyear[Theorem XX]{KLM}. The ability to similar
qualitative structures in $\PLclass^p$ breaks down when we examine infinite
algebra. This fact is a consequences of the results of \cite{FrHK1}.
}

Notice that \Pref{pro:default-iso} does not apply to
$\PLclass^{\epsilon}$.  Consider the PPD $(\Pr_1,\Pr_2, \ldots)$
such that $\Pr_n(a) = 1/n$, $\Pr_{2n-1}(b) = 1-1/n$, $\Pr_{2n-1}(c) =
0$, $\Pr_{2n}(b) = 0$, $\Pr_{2n}(c) = 1-1/n$ for all $n \ge 1$.
It is easy to check that the plausibility space arising from this
PPD is default-isomorphic to the one in \Pref{pro:order-iso}.

It is not hard to construct a trivial plausibility structure that is not
default-isomorphic to any structure in $\PLclass^\epsilon$:  Consider
the trivial plausibility measure on $\{a\}$ such that $\Pl(\{a\})
=\bot$.  This cannot be default-isomorphic to a structure in
$\PLclass^\epsilon$, since if $\Pl'$ is a plausibility measure in such a
structure, we must have $\Pl'(\{a\}) = \top > \Pl'(\emptyset)$.  But
this is essentially all that can go wrong.
We say
that a plausibility space $(W,\Pl)$
(\resp plausibility structure $(W,\Pl,\pi)$)
is {\em normal\/} (following Lewis
\citeyear{Lewis73})
if $\Pl(W) > \bot$.
It is easy to see that all structures in $\PLclass^{\epsilon}$,
$\PLclass^{\kappa}$, and $\PLclass^{\em Poss}$ are normal.
\thm\label{thm:epsilon-equivalent}
If $\PL \in \PLclass^{QPL}$ is a
normal plausibility structure
for a countable language $\L$,
then there is a structure $\PL' \in
\PLclass^{\epsilon}$ 
that is
default-equivalent to $\PL$.
\ethm
\prf
See Appendix~\ref{prf:epsilon}.
\eprf

\cor
If $(W,\Pl,\pi)$ is a normal, qualitative plausibility structure for a
countable language $\L$,
then there
exists a structure $(W,\Pl',\pi) \in \PLclass^{\epsilon}$ such that
$(W,\Pl,\pi)
\sat \phi\Cond\psi$ if and only if $(W,\Pl',\pi) \sat \phi\Cond\psi$
for all $\phi, \psi \in \L$ .
\ecor
Thus, with respect to conditional statements
in a countable language,
$\PLclass^{\epsilon}$ is as
    expressive (in a strong sense) as $\PLclass^{QPL}$.
However, for
    uncountable languages, there is a difference between
$\PLclass^{QPL}$ and $\PLclass^{\epsilon}$:
Probability distributions can assign positive weight only to a
countable number of pairwise disjoint events, while qualitative
plausibility measures do not suffer from such constraints.

\section{Epistemic Entrenchment}
\label{entrenchment}
\label{ENTRENCHMENT}

There has been much work related to defaults and plausibility.  It can
roughly be divided into three categories.  The first consists of
various approaches to dealing with uncertainty such as the ones
mentioned in \Sref{plausibility}. For a more detailed comparison to
such approaches see \cite{FrH7}. The second category consists of
semantics for defaults that are discussed at length in
Section~\ref{qualitative plausibility}.  The final
category includes semantics for defaults that are linguistic in
nature. The most well known approach of this kind is {\em epistemic
entrenchment\/} \cite{MakGar,Grove}.  This has been proposed as a
semantics for {\em belief revision\/} \cite{Gardenfors1}.  Recently,
G\"ardenfors and Makinson \citeyear{MakGar:Nonmon} proposed using a
similar notion of expectation ordering as a semantics for default
reasoning.
We briefly review
their approach here.

Let $\L$ be some logical language that includes the usual
propositional connectives with a compact consequence relation
$\vdash_\L$ that satisfies the axioms of propositional logic. An {\em
expectation ordering\/} $\Ele$ on $\L$ is a
relation over formulas in $L$ that satisfies the following
requirements:
\begin{itemize}\denselist
 \item[E1.] $\Ele$ is transitive,
 \item[E2.] if $\vdash_\L \phi \rimp \psi$ then $\phi \Ele \psi$,
 \item[E3.] for any $\phi$ and $\psi$, either $\phi \Ele \phi\land\psi$ or
    $\psi \Ele \phi\land\psi$.
\end{itemize}
Intuitively, $\phi \Ele \psi$ if $\psi$ is as at least as expected as
$\phi$, so the agent would not retract his belief in $\psi$ before
retracting his belief in $\phi$. We do not go here into the motivation
for E1--E3. It is not hard to verify that E1--E3 imply that $\Ele$ is
a total preorder on $\L$.

An {\em expectation structure\/} is a pair $E = (\L,\Ele)$, where
$\Ele$ is an expectation ordering on $\L$. Intuitively, $E$ satisfies
$\phi\Cond\psi$ if $\psi$ is the consequence of formulas that are
expected given $\phi$. This definition hinges on the choice of
formulas that are expected given $\phi$. G\"ardenfors and Makinson
take these to be the formulas that are more expected than $\neg\phi$.
Formally, an expectation structure $E=(\L, \Ele)$ satisfies a default
$\phi\Cond\psi$ if $\{ \phi \} \union \{ \xi : \neg\phi \Elt \xi \}
\vdash_\L \psi$, where $\phi \Elt \psi$ holds if $\phi \Ele \psi$ and
not $\psi \Ele \phi$.
The following result is almost immediate from the definitions:
\thm{\rm \cite{MakGar:Nonmon}}\label{thm:MakGar}
Let $E = (\L,\Ele)$ be an expectation structure. $E \sat
\phi\Cond\psi$ if and only if $\vdash_\L \phi \rimp \psi$ or
$(\phi\rimp\neg\psi) \Elt (\phi\rimp\psi)$.
\ethm

While this definition seems quite different than the one described in
\Sref{qualitative plausibility}, the two are in fact closely related.
Notice that $\phi \rimp \neg \psi$ is equivalent to $\neg(\phi \land
\psi)$, while $\phi \rimp \psi$ is equivalent to $\neg(\phi \land \neg
\psi)$.  Thus, the second clause in the theorem above, which says
$(\phi \rimp \neg \psi) \Elt (\phi \rimp \psi)$, can be viewed as
saying $\neg(\phi \land \psi)$ is less expected than $\neg(\phi \land
\neg \psi)$; this is clearly much in the spirit of the second clause
in the definition of defaults in plausibility, which says that $(\phi
\land\psi)$ must be more plausible than $(\phi \land \neg \psi)$.  If
we identify $p$ being more plausible than $q$ with $\neg p$ being less
expected than $\neg q$, they are equivalent.  The first clause in the
theorem, $\vdash_L \phi \rimp \psi$, corresponds to the vacuous case
that $\phi$ has plausibility $\bot$ in our definition.  However, as we
now show, G\"ardenfors and Makinson treat the vacuous case in a
somewhat nonstandard manner (which can still be captured using
plausibility).

To make the relationship between expectation orderings and plausibility
precise,
    let $E = (\L,\Ele)$ be an
    expectation structure. We say that a set $V \subseteq \L$ is {\em
    consistent\/} if for all $\phi_1, \ldots, \phi_n \in V$, we have
    $\not\vdash_\L \neg(\phi_1\land\ldots\land\phi_n)$.  $V$ is a
    {\em maximal consistent\/} set if it is consistent and for each $\phi \in
    \L$, either $\phi \in V$ or $\neg\phi\in V$. We now construct a
    plausibility structure $\PL_E = (W_E,
    \Pl_E, \pi_E)$. We define $W_E = \{ w_V : V $ is a maximally consistent
    subset of $\L \}$ and $\pi_E(w_V)(p) =$ {\bf true} if $p \in V$.
    Finally, we need to define $\Pl_E$.
    The obvious choice is to define
    $\Pl_E(\intension{\phi}) \le
    \Pl_E(\intension{\psi})$ if and only if $(\neg\psi) \Ele (\neg\phi)$.
It is easy to see that this implies that $(\phi\rimp\neg\phi) \Elt
    (\phi\rimp\psi)$ if
    and only if
    $\Pl_E(\intension{\phi\land\psi}) >
    \Pl_E(\intension{\phi\land\neg\psi})$. Thus, $E$ and $\PL_E$ agree
    on non-vacuous defaults.
However, suppose that $\Pl_E(\intension{\phi}) = \bot$ for some
consistent formula $\phi$.  It follows
that $\PL_E \sat \phi \Cond \psi$ for any $\psi$.  On  the other
hand, it is not hard to show that $E \sat \phi \Cond \psi$ if and
only if $\vdash_L \phi \rimp \psi$.%
\footnote{Proof sketch: The ``if'' direction follows from
Theorem~\ref{thm:MakGar}.
For the ``only if'' direction,
$\Pl_E(\intension{\phi}) = \bot$, so we must have
$\True \Ele \neg \phi$.  Since $\vdash_L \neg \phi \rimp
(\phi \rimp \neg \psi)$,
it follows from E1 and E2 that $\True \Ele (\phi \rimp
\neg \psi)$. Moreover, since $\vdash_L (\phi \rimp\psi) \rimp \True$
we have $(\phi\rimp\psi) \Ele \True$.
Thus,
we cannot have $(\phi \rimp \neg \psi) \Elt (\phi \rimp \psi)$, for
otherwise, by E2, we would have $\True \Elt \True$, a contradiction.}
We can easily
    modify the definition of $\Pl_E$ to avoid this problem:  We define
    $\Pl_E(\intension{\phi}) \le
    \Pl_E(\intension{\psi})$ either if
$(\neg\psi) \Ele  (\neg\phi)$ and it is not the case that
$\True \Ele \neg \psi$ or if
$\intension{\phi} \subseteq \intension{\psi}$.
With this modified definition, we get the desired result.
\pro\label{pro:epistemic}
If $E$ is an expectation structure, then $\PL_E$ is a plausibility
structure. Furthermore, $E \sat \phi\Cond\psi$ if
    and only if $\PL_E \sat \phi\Cond\psi$.
\epro
\prf
See Appendix~\ref{prf:entrenchment}.
\eprf

We now examine default entailment with respect to expectation orderings.
Let $\PLclass^E$ be the set of plausibility structures that correspond to
 expectation structures. It is not hard to prove that
\thm\label{thm:PE}
$\PLclass^E$ is a subset of $\PLclass^{QPL}$.
\ethm
\prf
It is straightforward to verify that if $E$ satisfies E1--E3, then
$\PL_E$ satisfies A2 and A3.
\eprf

It immediately follows that the KLM properties are sound for default
entailment with expectation orderings, \ie with respect to $\PLclass^E$.
The KLM properties, however, are
not complete with respect to $\PLclass^{E}$.  For example, if $p$ and $q$
are arbitrary primitive propositions, then $p \Cond \False$ entails $q
\Cond \False$.
\commentout{
This follows since the construction of $\PL_E$ implies
that $\Pl_E(A) = \bottom$ if and only if $A = \emptyset$. Thus, $\PLclass^E$
is not rich.  In particular, $\PLclass^E$ satisfies the following property,
called {\em consistency preservation\/} in \cite{MakGar:Nonmon}:
\begin{itemize}\denselist
 \item $\phi \Cond \False$ if and only if $\vdash_\L \phi \rimp \False$.
\end{itemize}
This property states that $\phi$ is totally {\em implausible \/}, \ie
has plausibility $\bot$, if and only if it is inconsistent. Thus, one
cannot specify that events, such as {\it
white}$\land${\it black}, are impossible in the database of $\Delta$. These
constraints must be somehow embedded in $\vdashL$.
}
This example is a consequence of a property G\"ardenfors and Makinson
call {\em consistency preservation\/}:
\begin{itemize}\denselist
 \item $E \sat \phi \Cond \False$ if and only if $\vdash_\L \phi \rimp \False$.
\end{itemize}
This property states that $\phi$ is totally {\em implausible\/}---that is,
has plausibility $\bot$---if and only if it is inconsistent.
This implies that no $\Pl_E \in \PLclass^E$ satisfies $p
\Cond\False$, and hence $p\Cond\False$ entails, vacuously, all other defaults.
Thus, one cannot specify that events such as {\it
white}$\land${\it black} are impossible in the database $\Delta$;
these constraints must be somehow embedded in $\vdashL$.

We note that expectation orderings are similar to plausibility
measures in that they order events. However, there are several
differences. First, expectation orderings use formulas to denote
events.  (We remark that there are similar formulations of probability
theory \cite{Jeffreys} that are based on a linguistic description of
events.)  Secondly, as shown by our construction of $\PL_E$,
expectation orderings order events according to the implausibility of
their complements. (This type of ordering is usually called the {\em
dual order\/} \cite{DuboisPrade88,FrH7,Shaf}.)  Thirdly, the treatment
of the vacuous case is slightly different.  This difference leads to
additional properties of default entailment.

\section{Conditional Logic}
\label{conditionallogic}
\label{CONDITIONALLOGIC}

Up to now, we focused on whether a set of defaults implies
another default.  We have not considered a full logic of defaults,
with negated defaults, nested defaults, and disjunctions of defaults.
It is easy to extend all the approaches we defined so far to deal
with such a logic.  {\em Conditional logic\/} is a logic that treats
$\Cond$ as a modal operator. The syntax of the logic is simple: let
$\LCond$ be the language defined by starting with primitive
propositions, and closing off under $\land$, $\neg$, and $\Cond$.
Formulas can describe logical combination of defaults (\eg $(p\Cond
q)\lor(p\Cond\neg q)$) as well as nested defaults (\eg $(p \Cond
q)\Cond r$).

We note that the connections between default reasoning and conditional
logics are well-known; see
\cite{Boutilier94AIJ1,CroccoLamarre,KLM,KatsunoSatoh}.
We gloss over the subtle philosophical differences between the two here.

The semantics of conditional logic is similar to the semantics of
defaults.
As with defaults, we evaluate conditional statements such as $\phi
\Cond \psi$ by comparing the plausibility of those worlds that satisfy
$\phi\land\psi$ to the plausibility of those worlds that satisfy $\phi
\land\neg\psi$. Unlike default reasoning, conditional logic allows us
to combine defaults with propositional statements. Thus,
$p \land (q \Cond r)$ is a formula of conditional logic, and is
satisfied if both $p$  and $q \Cond
r$ are satisfied. The truth of a formula such as $p \land (q \Cond r)$
depends on the world;  $p \land (q \Cond r)$ might be
true in $w_1$ and false in $w_2$
if $p$ is true at $w_1$ and not at $w_2$.

Conditional logic also allows us to consider nested
conditionals. For example, to evaluate $(p \Cond q) \Cond r$, we need
to consider the plausibility of the worlds that satisfy $r$ and $p
\Cond q$ and compare them to the plausibility of worlds that satisfy
$\neg r$ and $p \Cond q$. In the structures we considered in the
preceding sections, a statement such as $p \Cond q$ is determined by
the global plausibility measure. Thus, the set of worlds that satisfy
$p \Cond q$ is either the empty set or $W$ (\ie all possible worlds).
It is not hard to show that, as a result of this, we can {\em denest\/}
nested conditional statements.  That is, every formula is equivalent to
one without nested conditionals. (See \cite{FrH3} for a proof of this
well-known observation.)
 The usual definition of conditional logic \cite{Lewis73}
provides a
nontrivial semantics for nested conditionals by associating with each
world a different preferential order over worlds.
We can give a similar definition based on plausibility measures.

A {\em (generalized) plausibility structure\/} is a tuple
$(\Omega,\Plass,\pi)$ where $W$ and $\pi$ are, as usual, a set of worlds and
a mapping from worlds to truth assignments, and $\Plass$ maps each
world $w$ to a plausibility space $(W_w,\F_w,\Pl_w)$ where $W_w
\subseteq W$. Intuitively, $(W_w,\F_w,\Pl_w)$ describes the agent's
plausibility when she is in world $w$. We can view the plausibility
structures we defined in previous sections to be a special case
of generalized plausibility structures where $\Plass(w)$ is
the same for all worlds $w$.
For the remainder of this section we focus on generalized
plausibility structures, but continue to refer to them as
plausibility structures.

Given a plausibility structure $\PL = (\Omega,\Plass,\pi)$,
we define what it means for a formula $\phi$ to be true at a world $w$
in $\PL$.  The definition for the propositional connectives is
standard; for $\Cond$, we use the definition already given:
\begin{itemize}
\item $(\PL,w) \sat p$ if $\pi(w) \sat p$ for a primitive proposition
$p$
\item $(\PL,w) \sat \neg \phi$ if $(\PL,w) \not\sat \phi$
\item $(\PL,w) \sat \phi \land \psi$ if $(\PL,w) \sat \phi$ and
$(\PL,w) \sat \psi$
\item $(\PL,w) \sat \phi\Cond\psi$ if either
    $\Pl_w(\intension{\phi}_{(\sPL,w)}) = \bottom$ or
    $\Pl_w(\intension{\phi\land\psi}_{(\sPL,w)}) >
    \Pl_w(\intension{\phi\land\neg\psi}_{(\sPL,w)})$, where we define
    $\intension{\phi}_{(\sPL,w)} = \{ w\in W_w : (\Pl,w) \sat \phi \}$.%
\commentout{
\footnote{We redefine $\intension{\phi}_\sPL$ since $\phi$ can involve
    conditional statements. Note that if $\phi$ does not
    contain occurrences of $\Cond$ then this definition is equivalent
    to the one we gave earlier. Again, we omit the subscript when it is
    clear from the context.}
}
\end{itemize}

We can similarly define
generalized structures that use preferential
orderings, $\kappa$-rankings, $\epsilon$-semantics, or possibility
measures instead of plausibility measures. As before, all of these
structures can be embedded in qualitative plausibility  structures. We
denote by $\PLclass^{QPL}_c$  the class of all qualitative
generalized plausibility structures, and similarly denote the
subclasses that correspond to various semantics (\eg $\PLclass^p_c$
is the class that consists of plausibility structures based on
preference orderings).

We saw that with default reasoning, we could not distinguish between
plausibility, possibility, preference orderings, $\epsilon$-semantics,
and $\kappa$-rankings.  What happens when we move to the richer language
of conditional logic?
As we shall see, this richer language 
allows us to make some finer distinctions.

We start by examining preferential structures.
There is a complete axiomatization for
conditional logic with respect to preferential structures
due to Burgess \citeyear{Burgess81} called
System {\bf C}, consisting of the following axioms and inference rules:
\begin{itemize}\denselist
 \item[C0.] All substitution instances of propositional tautologies
 \item[C1.] $\phi \Cond \phi$
 \item[C2.] $((\phi \Cond \psi_1)\land(\phi\Cond\psi_2)) \rimp
    (\phi\Cond(\psi_1\land\psi_2))$
 \item[C3.] $((\phi_1\Cond\psi)\land(\phi_2\Cond\psi)) \rimp
    ((\phi_1\lor\phi_2) \Cond\psi)$
 \item[C4.] $((\phi_1\Cond\phi_2)\land(\phi_1\Cond\psi)) \rimp
    ((\phi_1\land\phi_2) \Cond \psi)$
 \item[R1.] From $\phi$ and $\phi \rimp \psi$ infer $\psi$
 \item[RC1.] From $\phi \dimp \phi'$ infer $(\phi\Cond\psi) \rimp
(\phi'\Cond\psi)$
 \item[RC2.] From $\psi \rimp \psi'$ infer $(\phi\Cond\psi) \rimp
(\phi\Cond\psi')$
\end{itemize}
System {\bf C} can be viewed as a generalization of \SysP.  The richer
language lets us replace a rule like AND by the axiom C2.  Similarly,
C1, C3, C4, RC1, and RC2 are the analogues of REF, OR, CM, LLE, and RW,
respectively.  We need C0 and R1 to deal with
propositional reasoning.
\thm
{\rm \cite{Burgess81}}
\label{MPREF}
System {\bf C} is a sound and complete axiomatization of $\LCond$ with
respect to $\PLclass^p_c$.
\ethm

Since the axioms of \SysC\ are clearly valid in all the structures in
$\PLclass^{QPL}_c$ and $\PLclass^p_c \subseteq \PLclass^{QPL}_c$,
we immediately get the following:
\thm\label{SysCcomplete}
System {\bf C} is a sound and complete axiomatization of $\LCond$
with respect to
$\PLclass^{QPL}_c$.
\ethm

The proof of \Tref{MPREF} given by Burgess is quite complicated.
We can get a simpler direct proof of \Tref{SysCcomplete}, without going
through \Tref{MPREF}, by using standard techniques of modal logic.  We
provide the details in Appendix~\ref{prf:conditionallogic}.
Theorems~\ref{MPREF} and~\ref{SysCcomplete} show that,
even in the richer framework of conditional logic, we cannot
distinguish between preferential orders and plausibility, at least not
axiomatically.
What about the other approaches we have been considering?

Not surprisingly, conditional logic does allow us to distinguish
rational structures from arbitrary preferential structures, because now
we can express RM within the language, using the following axiom:
\begin{itemize}\denselist
 \item[C5.]
$\phi\Cond\psi \land
\neg(\phi\Cond\neg\xi) \rimp \phi\land\xi \Cond \psi$
\end{itemize}
Does C5 (together with \SysC) characterize $\PLclass^{r}_c$?
Almost, but not
quite.  We say that a plausibility measure $\Pl$ is {\em rational\/} if
it satisfies
the following two properties:
\begin{itemize}\denselist
 \item[{\bf A4.}] For all pairwise disjoint sets $A, B$ and $C$, if $\Pl(A) <
\Pl(B)$, then $\Pl(A) < \Pl(C)$ or $\Pl(C) < \Pl(B)$.
 \item[{\bf A5.}] For all pairwise disjoint sets $A, B$ and $C$, if $\Pl(A) <
\Pl(B \union C)$, then $\Pl(A) < \Pl(B)$ or $\Pl(A) < \Pl(C)$.
\end{itemize}
A4 says that ordering of disjoint sets is  modular. (Recall that an
ordering is modular if there are no three elements $x,y,z$ such that
$x > y$ and $z$ incomparable to both $x$ and $y$.)  A5 says that
the plausibility of $B \union C$ is essentially the maximum of the
plausibility of $B$ and $C$. Thus, $B \union C$ cannot be more
plausible than $A$, unless one of the components is more plausible
than $A$.

It is not hard to show that C5 is valid in, and only in, systems where
the plausibility ordering is rational.
\pro\label{pro:rational}
Let $\PLclass \subseteq \PLclass_c^{QPL}$. C5 is valid in $\PLclass$
if and only if all structures in $\PLclass$ are rational.
\epro
\prf
See Appendix~\ref{prf:conditionallogic}.
\eprf

It is easy to verify that rational preference orderings,
$\kappa$-rankings, and possibility measures are all classes of
rational structures. It immediately follows that C5 is valid in each of
$\PLclass_c^r$, $\PLclass_c^{Poss}$, and $\PLclass_c^{\kappa}$.  On
the other hand, C5 it is not valid in $\PLclass_c^p$, since we can
easily construct preferential structures that violate A4 and A5.

As we said above, condition A4 states that the ordering is ``almost''
modular, in the sense that, when restricted to pairwise disjoint sets,
it is modular. It is not surprising to see modularity arise in this
context.  It is well known that a modular ordering induces a total
order. More precisely, if $<$ is a modular order on some set and we
define $x \le y$ as $y \not< x$, then $\le$ is a {\em total\/} order,
that is, a partial order such that either $x
\le y$ or $y \le x$ for all $x$ and $y$.
All approaches that satisfy rational monotonicity that have been proposed
in the literature involve structures where there is a total or modular
order on worlds (\eg rational preference orderings, $\kappa$-rankings,
and possibility measures).

We say that a plausibility measure $\Pl$ is a {\em ranking\/} if it satisfies
the following two properties:
\begin{itemize}\denselist
 \item[{\bf A4$'$.}] $\le_D$ is a total order; that is, either $\Pl(A)
\le_D \Pl(B)$ or $\Pl(B) \le_D \Pl(A)$ for all sets $A, B$.
 \item[{\bf A5$'$.}] $\Pl(A \union B) = \max(\Pl(A),\Pl(B))$ for all sets $A, B$.
\end{itemize}
It is easy to see that A4$'$ implies A4, and that in the presence of
A4$'$, A5$'$ implies A5 (A4$'$ is required to ensure that the two
plausibilities values have a maximum). Thus, any ranking is a
rational measure.
The opposite, however, is not true. It is easy to verify that A4 and
A5 do not imply A4$'$ and A5$'$. Thus, there is a discrepancy between the
properties that are necessary to satisfy C5 and those studied in the
literature. However, we now show that if we care only about defaults,
then there is no difference between rational structures and {\em  ranked
(plausibility) structures\/}, where the
plausibility measure is a ranking.

\thm\label{thm:Ranked}
If $(W,\Pl)$ be a rational qualitative plausibility space, then there is
a default-equivalent plausibility space $(W,\Pl')$ such that $\Pl'$ is a
ranking.
\ethm
\prf
See Appendix~\ref{prf:conditionallogic}.
\eprf

\cor\label{cor:Ranked}
If $\PL = (W,\P,\pi)$ is a rational plausibility structure, then
there is a ranked plausibility structure $\PL' = (W,\P',\pi)$ such that
$(\PL,w) \sat \phi$ if and only if $(\PL',w) \sat \phi$ for all worlds
$w$ and formulas $\phi \in \LCond$.
\ecor

\commentout{
In Corollary~\ref{cor:Ranked}, we do not say that the ranked structure
was qualitative.  In fact,
rankings are always qualitative:
\pro\label{pro:Rational-Qual}
Let $\Pl$ be a plausibility measure. If $\Pl$ satisfies A1, A4$'$ and
A5$'$, then it also satisfies A2 and A3.
\epro
\prf
See Appendix~\ref{prf:conditionallogic}.
\eprf
}

Conditional logic allows us to capture another property that we 
encountered earlier. Recall that measures based on PPDs, possibility
measures, and $\kappa$-rankings are all {\em normal\/},
that is, that $\Pl(W) > \bot$. This property corresponds to the
axiom

\begin{itemize}\denselist
 \item[C6.] $\neg(\true\Cond\false)$.
\end{itemize}
It is not hard to show that C6 is
valid in each of $\PLclass_c^{Poss}, \PLclass_c^{\kappa}$, and
$\PLclass^\epsilon$.

Using C5 and C6  we can characterize $\PLclass_c^\epsilon$,
$\PLclass_c^\kappa$, and $\PLclass_c^{Poss}$.

\thm\label{thm:AX-all}
\begin{enumerate}\denselist
 \item[(a)] {\bf C}$+\{$C6$\}$ is a sound and complete axiomatization
of $\PLclass_c^\epsilon$.
 \item[(b)] {\bf C} $+ \{$C5, C6$\}$ is a sound and complete
axiomatization of $\PLclass_c^\kappa$
and $\PLclass_c^{Poss}$.
\end{enumerate}
\ethm
\prf
See Appendix~\ref{prf:conditionallogic}.
\eprf

\section{Conclusions}\label{discussion}

We feel that this paper makes three major contributions: the
introduction of plausibility measures, the unification of all earlier
results regarding the KLM properties into one framework, and a general
result showing the inevitability of these properties.

Do we really
need plausibility measures?
That depends on the language we are interested in.
If all we are interested in is
propositional default reasoning and the KLM properties,
then, as is well known (and our results emphasize), many different
approaches turn out to be equivalent in expressive power.
If we move to the richer language of propositional conditional logic,
then, as the results of Section~\ref{conditionallogic} show, we start to
see some differences (that are captured by axioms C5 and C6, which
correspond to rationality and normality, respectively), although
plausibility structures  and preferential structures continue to be
characterized by the same axioms.  As we show in a companion paper
\cite{FrHK1}, once we move to first-order conditional logic, more
significant differences start to appear.
The extra expressive power of plausibility structures makes them more
appropriate than preferential structures for providing semantics for
first-order default reasoning.  This difference is due to
the fact when doing propositional reasoning, we
can safely restrict to finite structures.  (Technically, this is
because we have a {\em finite model property}: if a formula in
$\LCond$ is satisfiable, it is satisfiable in a finite plausibility
structure; see Lemma~\ref{lem:finite-conditional-structure}.)
In finite structures, preferential orders and plausibility measures are
equi-expressive.
The differences that we observe between them in first-order conditional
logic are due to the fact that in first-order reasoning,
infinite structures play a more important role.

Beyond their role in default reasoning, we expect that plausibility
measures will prove useful whenever we want to express uncertainty and
do not want to (or cannot) do so using probability.  For example, we
can easily define a plausibilistic analogue of conditioning
\cite{FrH7}.  While this can also be done in many of the other
approaches we have considered, we believe that the generality of
plausibility structures will allow us to again see what properties of
independence we need for various tasks.  In particular, in
\cite{FrH6}, we use plausibilistic independence to define a
plausibilistic analogue of Markov chains.  In future work we plan to
explore further the properties and applications of plausibility
structures.

\begin{acks}
The authors are grateful to Ronen Brafman, Adnan Darwiche, Moises
Goldszmidt, Adam Grove, Daphne Koller, Daniel Lehmann, Karl Schlechta,
and Zohar Yakhini  for useful
discussions
relating to this work.
\commentout{
The authors were supported in part by the Air
Force Office of Scientific Research (AFSC), under Contract
F49620-91-C-0080 and by NSF grant IRI-95-03109.  The first author
was also supported in part by
IBM Graduate Fellowship and
Rockwell Science Center.
}
\end{acks}

\appendix
\section{Detailed Proofs}\label{app:proofs}
\label{app:plaus-proofs}

\subsection{Proofs for Section~\protect\ref{qualitative plausibility}}
\label{prf:qualplaus}

\relem{lem:embedding}
Let $W$ be a set of possible worlds and let $\pi$ be a function that maps each
world in $W$ to a truth assignment to $\L$.
Let $T \subseteq \LDef$ be a set of defaults that is closed under the
rules of \SysP\ that satisfies the following condition:
\begin{itemize}
\item[{\rm ($*$)}]\it if $\phi \Cond \psi \in T$, $\intension{\phi} =
\intension{\phi'}$, and
$\intension{\psi} = \intension{\psi'}$,  then $\phi' \Cond \psi' \in T$, for
all formulas $\phi, \phi', \psi, \psi' \in \L$.
\end{itemize}
There is a plausibility structure $\PL_T = (W,\Pl_T, \pi)$ such that
$\Pl_T(\intension{\phi}) \le \Pl_T(\intension{\psi})$ if and
only if $\phi\lor\psi \Cond \psi \in T$.
Moreover,
$\PL_T \sat \phi \Cond \psi$ if and only if $\phi \Cond \psi \in T$.
\erelem

\prf
We start by noting that the condition imposed on $T$ ensures that
$\Pl_T$ is well defined, \ie sets that are described by different
formulas are compared in a consistent manner.
We now examine whether there exists a plausibility measure $(W,
\Pl_T)$ satisfying ($*$). It suffices
to show that the order relation on $\Pl_T$ is reflexive, transitive, and
satisfies A1.

\begin{description}\denselist
 \item[Reflexivity]  Applying REF and LLE, we get that $(\phi\lor\phi)
\Cond \phi \in T$ for all $\phi$.
 \item[Transitivity] This is a direct consequence of the following
lemma of Kraus, Lehmann and Magidor.
\lem{\rm \cite[Lemma 5.5]{KLM}}
Let $T$ be a set of defaults closed under applications of the rules of
\SysP.
Then if both
$\phi_1 \lor \phi_2 \Cond \phi_2$ and $\phi_2 \lor \phi_3
\Cond \phi_3$ are in $T$, then so is $\phi_1 \lor \phi_3 \Cond \phi_3$.
\elem
\item[A1] Assume $\intension{\phi} \subseteq \intension{\psi}$.
It follows that $\intension{\psi} =
\intension{\psi\lor\phi}$.
Since $T$ is closed under REF we get that $(\phi \lor \psi) \Cond
(\phi\lor\psi) \in T$.
Using ($*$), we get that $(\phi \lor \psi) \Cond
\psi \in T$. Thus,
$\Pl_T(\intension{\phi}) \le \Pl_T(\intension{\psi})$.
\end{description}

Finally, we need to show that $(W,\Pl_T,\pi) \sat \phi \Cond \psi$ if
and only if $\phi\Cond\psi \in T$. We start by observing that LLE,
REF, and AND imply that $\phi\Cond\psi \in T$ if and only if
$(\phi\land\neg\psi)\lor(\phi\land\psi) \Cond \phi\land\psi \in T$.
We conclude that
\begin{quote}
{\rm ($**$)}
$\phi\Cond\psi \in T$ if and only if
$\Pl_T(\intension{\phi\land\neg\psi}) \le
\Pl_T(\intension{\phi\land\psi})$.
\end{quote}
Thus, it suffices to show that
$\Pl_T(\intension{\phi\land\neg\psi}) \le
\Pl_T(\intension{\phi\land\psi})$ if and only if
either $\Pl_T(\intension{\phi\land\neg\psi}) <
\Pl_T(\intension{\phi\land\psi})$ or $\Pl_T(\intension{\phi}) = \bot$.
The ``if'' direction is trivial. For the ``only if'' direction, suppose
by way of contradiction that we have $\Pl_T(\intension{\phi}) > \bot$,
 $\Pl_T(\intension{\phi\land\neg\psi}) \le
\Pl_T(\intension{\phi\land\psi})$,
and
$\Pl_T(\intension{\phi\land\psi}) \le
\Pl_T(\intension{\phi\land\neg\psi})$. From ($**$), we have that $\phi
\Cond \psi \in T$ and $\phi\Cond\neg\psi \in T$. By AND and ($*$), we
have that $\phi\Cond\False \in T$. But then $\Pl_T(\intension{\phi}) =
\bot$, which contradicts our assumptions.
\commentout{
Assume that $\phi \Cond\psi \in T$. We just showed that this implies that
$\Pl_T(\intension{\phi\land\neg\psi}) \le
\Pl_T(\intension{\phi\land\psi})$. If
$\Pl_T(\intension{\phi\land\psi}) \not\le
\Pl_T(\intension{\phi\land\neg\psi})$, then $\PL \sat \phi\Cond\psi$.
So assume that $\Pl_T(\intension{\phi\land\psi}) \le
\Pl_T(\intension{\phi\land\neg\psi})$. By definition, we have that
$(\phi \land \neg\psi \lor \phi\land\psi)  \Cond \phi\land\neg \psi
\in T$. We also have that $(\phi \land \neg\psi \lor \phi\land\psi)
\Cond \phi\land \psi \in T$. By using the AND and ($*$) we get
that $(\phi \lor \False) \Cond \False \in T$. Thus,
$\Pl_T(\intension{\phi}) \le \bot$, $\PL_T \sat \phi\Cond\psi$.

Assume that $(W, \Pl_T,\pi) \sat \phi\Cond\psi$. There are two
possible cases.
\commentout{
First assume that $\Pl_T(\intension{\phi}) = \bot$.
Then, by definition, $(\phi \lor \False) \Cond \False \in T$. Applying
LLE and RW we
get that $\phi \Cond \psi \in T$. Now assume that
$\Pl_T(\intension{\phi\land\psi}) >
\Pl_T(\intension{\phi\land\neg\psi})$. Thus,
$\phi\land\psi \lor \phi\land\neg\psi \Cond \phi\land\psi \in T$. We
conclude that $\phi \Cond\psi \in T$.
}
If $\Pl_T(\intension{\phi}) = \bot$, then by A1 we get that
$\Pl_T(\intension{\phi\land\neg\psi}) =
\Pl_T(\intension{\phi\land\psi}) = \bot$. Otherwise, we get that
$\Pl_T(\intension{\phi\land\neg\psi}) <
\Pl_T(\intension{\phi\land\psi})$. In both cases
$\Pl_T(\intension{\phi\land\neg\psi}) \le
\Pl_T(\intension{\phi\land\psi})$. As we showed above, this implies
that $\phi \Cond \psi \in T$.
}
\eprf

\subsection{Proofs for Section~\protect\ref{axioms}}
\label{prf:axioms}

\rethm{thm:SysP'}
If $\Delta \vdashpp \phi\Cond\psi$, then $\Delta \sat_{\PLclass^{PL}} \phi\Cond\psi$.
\erethm

\prf
We need to show that LLE, RW, and REF are sound in $\PLclass^{\PL}$. Let
$\PL = (W,\Pl,\pi)$ be a plausibility structure. We proceed as follows.
\begin{description}\denselist
 \item[LLE] Assume that $\vdashL \phi \dimp \phi'$. Then, by definition,
$\intension{\phi} = \intension{\phi'}$. The
soundness of LLE immediately follows.

 \item[RW]  Assume that $\vdashL \psi \rimp \psi'$ and that $\PL \sat \phi
\Cond \psi$. We want to show that $\PL \sat \phi\Cond\psi'$. If
$\Pl(\intension{\phi}) = \bot$, this is immediate. On the other hand,
if $\Pl(\intension{\phi}) > \bot$, then
$\Pl(\intension{\phi\land\neg\psi}) < \Pl(\intension{\phi\land\psi})$.
Since $\vdashL \psi\rimp\psi'$ we have that
$\intension{\psi} \subseteq \intension{\psi'}$. It follows that
$\intension{\phi \land\neg \psi'} \subseteq
\intension{\phi\land\neg\psi}$ and
$\intension{\phi \land \psi} \subseteq \intension{\phi\land\psi'}$.
Using A1, we conclude that
$\Pl(\intension{\phi\land\psi'} >
\Pl(\intension{\phi\land\neg\psi'}$, so $\PL \sat \phi \Cond\psi'$.

 \item[REF]
By definition, $\intension{\phi \land \neg \phi} = \emptyset$ and
$\Pl(\emptyset) = \bot$. Thus, if
$\Pl(\intension{\phi}) > \bot$, then $\PL \sat \phi \Cond\phi$. On the
other hand, if $\Pl(\intension{\phi}) = \bot$, then $\PL \sat
\phi\Cond\phi$ vacuously.
\end{description}
\eprf

\repro{pro:A2} A plausibility measure satisfies A2 if and only if it satisfies A2$'$.
\erepro

\prf
Let $(W,\Pl)$ be a plausibility space. Assume that  $\Pl$ satisfies A2.
Let $A$, $B_1$, and $B_2$ be sets such that
$\Pl(A \inter B_1) > \Pl(A \inter \overline{B_1})$ and $\Pl(A \inter
B_2) > \Pl(A \inter \overline{B_2})$. Set $C = A \inter B_1 \inter
B_2$, $D = A \inter B_1 \inter \overline{B_2}$, and $E = A \inter
\overline{B_1}$. It is easy to verify that $C$, $D$, and $E$ are pairwise
disjoint. Since $C \union D = A \inter B_1$, we have that
$\Pl( C \union D ) > \Pl(E)$. Moreover, since $C \union E \supseteq A
\inter B_2$, $D \subseteq A \inter \overline{B_2}$, and
$\Pl(A \inter B_2) > \Pl(A \inter \overline{B_2})$,
we can apply A1
and conclude that $\Pl(C \union E ) > \Pl(D)$. Applying A2, we conclude
that $\Pl(C) > \Pl(D \union E)$, \ie $\Pl(A \inter B_1 \inter B_2) >
\Pl(A \inter \overline{(B_1 \inter B_2)})$. This gives us A2$'$.

Now assume that $\Pl$ satisfies A2$'$. Let $C, D$, and $E$ be pairwise
disjoint sets such that $\Pl(C \union D) > \Pl(E)$ and $\Pl(C \union
E) > \Pl (D)$. Let $A = C \union D \union
E$, $B_1 = C\union D$, and $B_2 = C \union E$. Then we have $\Pl(A
\inter B_1) = \Pl(C \union D) > \Pl(E) = \Pl(A \inter \overline{B_1})$
and
$\Pl(A \inter B_2) = \Pl(C \union E) > \Pl(D) = \Pl(A \inter
\overline{B_2})$.
{From} A2$'$, we have that
$\Pl(A \inter B_1 \inter B_2) > \Pl(A \inter \overline{(B_1 \inter
B_2)})$, \ie $\Pl(C) > \Pl(D \union E)$. This gives us A2.
\eprf

\rethm{QPL}
$\P \subseteq \PLclass^{QPL}$ if and only if  for
all $\Delta,\phi$, and $\psi$,
if $\Delta\vdashp
    \phi\Cond\psi$, then $\Delta\sat_\PLclass \phi\Cond\psi$.
\erethm

\prf
To prove the ``if'' direction it suffices to show that each rule in
\SysP\ is sound in qualitative structures.
Let $\PL = (W, \Pl, \pi)$ be a qualitative plausibility structure. The
soundness of LLE, RW, and REF is proved in \Tref{thm:SysP'}. To deal with the
remaining cases, we proceed as follows.
\begin{description}\denselist
 \item[AND] Assume that $\PL \sat \phi \Cond \psi_1$ and $\PL \sat \phi \Cond
\psi_2$. If $\Pl(\intension{\phi}) = \bot$, then $\PL \sat
\phi\Cond\psi_1\land\psi_2$ vacuously. Assume that
$\Pl(\intension{\phi}) > \bot$. Let $A = \intension{\phi}$, $B_1 =
\intension{\psi_1}$, and $B_2 = \intension{\psi_2}$.
Since $\PL \sat \phi\Cond\psi_1$ and $\PL \sat \phi\Cond\psi_2$, we
have that $\Pl(A \inter B_1) > \Pl(A
\inter \overline{B_1})$ and
$\Pl(A \inter B_2) > \Pl(A \inter \overline{B_2})$. \Pref{pro:A2} states
that $\Pl$ satisfies A2$'$, and thus we get that
$\Pl(A \inter B_1 \inter B_2) > \Pl(A \inter \overline{(B_1 \inter
B_2)})$, so  $\PL \sat \phi \Cond \psi_1\land\psi_2$.

 \item[CM] Again assume that $\PL \sat \phi\Cond\psi_1$ and $\PL \sat
\phi\Cond\psi_2$. If $\Pl(\intension{\phi \land \psi_1}) = \bot$, then
$\PL \sat \phi\land\psi_1\Cond\psi_2$ vacuously. Assume that
$\Pl(\intension{\phi\land\psi_1}) > \bot$. Let $A, B_1$, and $B_2$ be
defined as in the treatment of AND above. Again, we have $\Pl((A \inter B_1 \inter B_2) > \Pl(A \inter \overline{(B_1 \inter
B_2)})$.
Since $A \inter B_1 \inter
\overline{B_2} \subseteq A \inter \overline{(B_1 \inter
B_2)}$, we conclude that
$\Pl(A \inter B_1 \inter B_2) > \Pl(A \inter B_1 \inter
\overline{B_2})$. Thus, $\PL \sat \phi \land \psi_1 \Cond \psi_2$.

 \item[OR] Assume that $\PL \sat \phi_1\Cond\psi$ and $\PL \sat
\phi_2\Cond\psi$. If $\Pl(\intension{\phi_1}) =
\Pl(\intension{\phi_2}) = \bot$, then applying A3 we get that
$\Pl(\intension{\phi_1\lor\phi_2}) = \bot$ and thus $\PL \sat
(\phi_1\lor\phi_2) \Cond \psi$ vacuously. So  assume that
$\Pl(\intension{\phi_1}) > \bot$. (Identical argument works if
$\Pl(\intension{\phi_2}) > \bot$.) Set $A = \intension{(\phi_1 \lor
\phi_2) \land \psi}$, $B = \intension{\phi_1 \land \neg \psi}$, and $C
= \intension{\phi_2 \land \neg \phi_1 \land \neg \psi}$. To prove that
$\PL \sat (\phi_1\lor\phi_2) \Cond \psi$, we must show that $\Pl(A) >
\Pl(B\union C)$. Since $\PL \sat \phi_1 \Cond \psi$, we have that
$\Pl(A) \ge \Pl(\intension{\phi_1 \land \psi}) > \Pl(B)$. If
$\Pl(\intension{\phi_2}) = \bot$, then
$\Pl(C) = \bot$ and we conclude that $\Pl(A) > \Pl(C)$. On
the other hand, if  $\Pl(\intension{\phi_2}) > \bot$, then since $\PL
\sat \phi_2 \Cond \psi$, we have $\Pl(A) \ge
\Pl(\intension{\phi_2 \land \psi}) >
\Pl(\intension{\phi_2 \land \neg \psi}) \ge \Pl(C)$. From A2, $\Pl(A)
> \Pl(B)$, and $\Pl(A) > \Pl(C)$, we
get that $\Pl(A) > \Pl(B \union C)$. Thus, $\PL \sat
(\phi_1 \lor \phi_2) \Cond \psi$.
\end{description}

To prove the ``only if'' direction we have to show that if there is
some $\PL = (W,\Pl,\pi)$ in $\PLclass$ that is not qualitative, then the KLM
properties are not sound with respect to $P$. Assume that $\Pl$ does
not satisfy A2. Since we have assumed that $\F = \{ \intension{\phi} :
\phi \in \L \}$, there are formulas $\phi$, $\psi_1$, and
$\psi_2$, such that $\intension{\phi}$, $\intension{\psi_1}$, and
$\intension{\psi_2}$ are pairwise disjoint,
$\Pl(\intension{\phi\lor\psi_2}) > \Pl(\intension{\psi_1})$,
$\Pl(\intension{\phi\lor\psi_1}) > \Pl(\intension{\psi_2})$, and yet
$\Pl(\intension{\phi}) \not> \Pl(\intension{\psi_1\land\psi_2})$.
Thus, $\PL \sat (\phi\lor\psi_1\lor\psi_2) \Cond \neg\psi_1$ and
$\PL \sat (\phi\lor\psi_1\lor\psi_2) \Cond \neg\psi_2$. However, $\PL
\not\sat (\phi\lor\psi_1\lor\psi_2 )\Cond (\neg\psi_1 \land \neg \psi_2)$.
This shows that the AND rule is not sound in $\PLclass$.
Now assume that there is some $\PL = (W,\Pl,\pi)$ that does not satisfy
A3. Thus, there are formulas
$\phi_1$ and $\phi_2$ such that $\Pl(\intension{\phi_1}) =
\Pl(\intension{\phi_2}) = \bot$ and $\Pl(\intension{\phi_1\lor\phi_2})
> \bot$. We conclude that $\PL \sat\phi_1 \Cond \False$ and
$\PL \sat \phi_2 \Cond \False$, but $\PL \not\sat (\phi_1\lor\phi_2) \Cond
\False$. This shows that the OR rule is not sound in $\PLclass$.
\eprf

\rethm{everythinginQPL}
Each of
$\PLclass^{Poss}$, $\PLclass^\kappa$, $\PLclass^\epsilon$, $\PLclass^p$, and $\PLclass^r$ is a
subset of  $\PLclass^{QPL}$. 
\erethm
\prf
It is straightforward to verify that A2 and A3 hold for each structure
in $\PLclass^{Poss}$,
$\PLclass^\kappa$, $\PLclass^\epsilon$, $\PLclass^p$, and
$\PLclass^r$.

We start with $\PLclass^{Poss}$.  Let $(W, \Poss,
\pi)$ be a possibility structure. 
To prove A2, assume that $A, B, C
\subseteq W$ are pairwise disjoint sets such that $\Poss(A \union B) >
\Poss( C )$ and $\Poss(A \union C) > \Poss( B )$. Since $\Poss(A
\union B) = \max( \Poss(A), \Poss(B) )$, we have that $\max( \Poss(A),
\Poss(B) ) > \Poss(C)$ and that $\max(
\Poss(A), \Poss(C) ) > \Poss(B)$. It 
easily follows that $\Poss(A) >
\max(\Poss(B),\Poss(C)) = \Poss(B \union C)$, as required by A2.
To prove A3, 
suppose
that $\Poss(A)
= \Poss(B) = 0$. Since $\Poss( A \union B ) =
\max(\Poss(A),\Poss(B))$, we have that $\Poss(A \union B) = 0$, as
required by A3.

The proof for $\PLclass^{\kappa}$ is identical (replacing $\max$ and
$0$ with $\min$ and $\infty$, respectively).

Next, 
consider
$\PLclass^{\epsilon}$. Let $(W, \{ \Pr_i \},
\pi)$ be a PPD structure and let $(W, \Pl_{PP}, \pi)$ be
the corresponding structure in $\PLclass^{\epsilon}$. To prove A2,
assume that $A, B, C \subseteq W$ are pairwise disjoint sets such that
$\Pl_{PP}(A \union B) > \Pl_{PP}( C )$ and $\Pl_{PP}(A \union C) >
\Pl_{PP}( B )$. We want to show that $\Pl_{PP}(A) > \Pl_{PP}(B \union
C)$. According to the construction of
Theorem~\ref{thm:plausembedding}, we need to show that $\lim_{i
\rightarrow \infty} \Pr_i( A \mid A
\union B \union C ) = 1$ and that  
$\lim_{i \rightarrow \infty} \Pr_i(
B \union C \mid A \union B \union C ) \neq 1$ (the limit can be
undefined in this case). Since $\Pl_{PP}(A \union B) > \Pl_{PP}( C )$
and $\Pl_{PP}(A \union C) > \Pl_{PP}( B )$, we have that
\begin{eqnarray}
&\lim_{i \rightarrow \infty} {\Pr}_i( A \union B \mid A \union B \union
C ) = 1,\,
\lim_{i \rightarrow \infty} {\Pr}_i( A \union C \mid A \union B \union
C ) = 1 & \label{eq:eps-1}\\
&\lim_{i \rightarrow \infty} {\Pr}_i( B \mid A \union B  \union C ) \neq
1,\,
\lim_{i \rightarrow \infty} {\Pr}_i( C \mid A \union B  \union C ) \neq
1 & \label{eq:eps-2}
\end{eqnarray}
To prove that $\lim_{i\rightarrow \infty} \Pr_i( A \mid A
\union B \union C ) = 1$, fix $\epsilon > 0$. From (\ref{eq:eps-1}),
we have that there is an $n_\epsilon$ such that 
for all $i > n_\epsilon$, $\Pr_i(A \union B
\mid A \union B \union C ) > 1 - \frac{\epsilon}{2}$ and  $\Pr_i(A \union C
\mid A \union B \union C ) > 1 - \frac{\epsilon}{2}$. Let $i >
n_\epsilon$. There are two 
cases. If $\Pr_i(A \union B \union C) = 0$, then $\Pr_i(A \mid A
\union B \union C )  = 1$ by definition. If $\Pr_i(A \mid A \union B
\union C ) > 0$, we use the disjointness of $A$, $B$, and $C$ to get 
\begin{eqnarray*} 
{\Pr}_i( A  \mid A \union B \union C ) + {\Pr}_i( B \mid A \union B
\union C ) & > & 1 - \frac{\epsilon}{2} \\
{\Pr}_i( A \mid A \union B \union C ) + {\Pr}_i( C \mid A \union B
\union C ) & > & 1 - \frac{\epsilon}{2}
\end{eqnarray*}
This implies that $\Pr_i( A  \mid A \union B \union C ) + \Pr_i( A
\union B \union C \mid A \union B \union C ) > 2 - \epsilon$. Since $\Pr_i( A
\union B \union C \mid A \union B \union C ) = 1$, we get that
$\Pr_i( A \mid A \union B \union C ) > 1 - \epsilon$. We conclude that
$\Pr_i( A \mid A \union B \union C ) > 1 - \epsilon$ for all $i >
n_\epsilon$, and thus $\lim_{i\rightarrow \infty} \Pr_i( A \mid A
\union B \union C ) = 1$.

To prove that
$\lim_{i\rightarrow \infty} \Pr_i( B \union C \mid A \union B \union C
) \neq 1$, it suffices to find a subsequence on which $\Pr_i( B \union
C \mid A \union B \union C ) \rightarrow 0$. Let $i_1, i_2, \ldots,
i_j, \ldots$ be the sequence of indexes such that $\Pr_{i_j}(A \union B
\union C) > 0$. 
This sequence must be infinite, for otherwise, since $\Pr_i(B \mid A
\union B \union  
C) = 1$ whenever $\Pr_i(A \union B \union C) = 0$, we would have that
$\lim_{i \rightarrow \infty} \Pr_i( B \mid A \union B  \union C ) =
1$, contradicting (\ref{eq:eps-2}).
{From} $\lim_{i\rightarrow \infty} \Pr_i( A \mid A
\union B \union C ) = 1$, we have that $\lim_{j\rightarrow \infty}
\Pr_{i_j}( A \mid A \union B \union C ) = 1$. Moreover, since
$\Pr_{i_j}(B \union C \mid A \union B \union C) = 1 - \Pr_{i_j}(A \mid
A \union B \union C)$, we get that $\lim_{j\rightarrow \infty}
\Pr_{i_j}( B \union C \mid A \union B \union C ) = 0$. We conclude
that $\lim_{i\rightarrow \infty} \Pr_i( B \union C \mid A \union B
\union C ) \neq 1$.

To prove A3, assume that $A, B \subseteq W$ are such that $\Pl_{PP}(A)
= \Pl_{PP}(B) = \bot$. By the construction of
Theorem~\ref{thm:plausembedding}, we have that $\Pl_{PP}(A) \le
\bottom = 
\Pl_{PP}(\emptyset)$ if $\lim_{i \rightarrow \infty} \Pr_i(\emptyset
\mid A \union \emptyset ) = 1$. This implies that there is an index
$n_A$ such that $\Pr_i(A) = 0$ for all $i > n_A$. Similarly, there is
an $n_B$ such that $\Pr_i(B) = 0$ for all $i > n_B$. Hence, $\Pr_i(
A\union B ) = 0$ for all $i > \max(n_A, n_B)$.  We conclude that
$\Pl_{PP}(A \union B) = \bot$.

Finally, we 
consider $\PLclass^{p}$ and $\PLclass^{r}$. Let $(W, \prec,
\pi)$ be a preference structure and let $(W, \Pl_{\prec}, \pi)$ be the
corresponding structure in $\PLclass^{p}$.
To prove A2,
assume that $A, B, C \subseteq W$ are pairwise disjoint sets such that
$\Pl_{\prec}(A \union B) > \Pl_{\prec}( C )$ and $\Pl_{\prec}(A \union C) >
\Pl_{\prec}( B )$. We want to show that $\Pl_{\prec}(A) > \Pl_{\prec}(B \union
C)$. It is easy to verify that the construction of
Theorem~\ref{thm:plausembedding} is such that disjoint sets cannot
have equal plausibilities. Thus, it suffices to show that
$\Pl_{\prec}(A) \ge \Pl_{\prec}(B \union C)$.  That is, for all $w \in B
\union C$ there is a world $w' \in A$ such that (a) $w' \prec w$ and
(b) there is no $w'' \in B \union C$ such that $w'' \prec w'$.  Let
$w_{BC} \in B \union C$. 
Without loss of generality, we can assume that $w \in B$. 
Since $\Pl_{\prec}(A \union C) >
\Pl_{\prec}(B)$, there is a world $w_{AC} \in A \union C$ such that $w_{AC}
\prec w_{BC}$ and for all $w_B \in B$, $w_B \not\prec w_{AC}$. There are
three cases: (1)
If $w_{AC} \in C$, then since $\Pl_{\prec}(A \union B) >
\Pl_{\prec}(C)$, there is a world $w_{AB} \in A \union B$ such that $w_{AB} \prec w_{AC}$ and for all $w_C \in C$, $w_C \not
\prec w_{AB}$. Since $w_B \not\prec w_{AC}$ for all $w_B \in B$, we
get that $w_{AB} \in A$. For requirement (a), by 
the
transitivity of
$\prec$ we have that $w_{AB} \prec w_{BC}$. For requirement (b), 
suppose that
$w'' \in B \union C$. If $w'' \in C$, then we have that $w'' \not\prec
w_{AB}$. On the other hand, if $w'' \in B$, then we have that $w''
\not \prec w_{AC}$, and by transitivity $w'' \not\prec w_{AB}$.  
(2) If
$w_{AC} \in A$ and there is a world $w_C \in C$ such that $w_C \prec
w_{AC}$, then since $\Pl_{\prec}(A \union B) > \Pl_{\prec}(C)$, there
is a world $w_{AB}$ such that $w_{AB} \prec w_C$ and for all worlds
$w'' \in C$, $w'' \not
\prec w_{AB}$. Again, it follows that $w_{AC}
\in A$ and satisfies (a) and (b).
Finally, (3) if $w_{AC} \in A$ and for all $w_C \in C$, $w_C \not\prec
w_{AC}$, then it is easy to check that $w_{AC}$ satisfies (a) and (b).

To prove A3, we note that the construction of
Theorem~\ref{thm:plausembedding} is such that $\Pl_{\prec}(A) = \bot$
if and only if $A = \emptyset$. A3 immediately follows.
\eprf

\rethm{thm:main}
A set $\PLclass$ of qualitative plausibility structures is rich if and
only if for all finite $\Delta$ and defaults $\phi \Cond \psi$, we
have that $\Delta\sat_\PLclass\phi\Cond\psi$ implies
$\Delta\vdashp\phi\Cond\psi$.
\erethm
\prf
For the ``if'' direction, assume that $\PLclass$ is not rich.
We need to show that \SysP\ is not complete for $\sat_\PLclass$. It is
sufficient to construct $\Delta$, $\phi$, and $\psi$ such that $\Delta
\sat_\PLclass \phi \Cond\psi$ but $\Delta \not\vdashp \phi\Cond\psi$.

We start with a lemma, whose straightforward proof is left to the reader.
\lem\label{lem:Delta}
Let $\phi_1, \ldots, \phi_n$ be a collection of mutually exclusive
formulas. Let $\Delta$ consist of the default $\phi_{n} \Cond\False$
and the defaults $\phi_{i} \lor \phi_{j} \Cond \phi_{i}$ for all $1
\le i < j \le n$.  Then $(W, \Pl, \pi) \sat \Delta$ if and only if
there is some $j$ with $1 \le j \le n$ such that $$
\Pl(\intension{\phi_1}) > \Pl( \intension{\phi_2}) >
\cdots > \Pl(\intension{\phi_j}) = \cdots = \Pl(\intension{\phi_n}) = \bot.
$$
\elem

Since $\PLclass$ is not rich, there is a collection $\phi_1, \ldots, \phi_n$
    that is a counterexample to the definition of richness.
Let $\Delta$ be the set of defaults defined in \Lref{lem:Delta}. We
claim that if $(W,\Pl,\pi) \in \PLclass$ satisfies all the defaults in
$\Delta$, then $\Pl(\intension{\phi_{n-1}}) = \bot$. To see this,
assume that $\Pl(\intension{\phi_{n-1}}) > \bot$. Then according to
\Lref{lem:Delta}, $\Pl(\intension{\phi_1}) >
\cdots > \Pl(\intension{\phi_{n-1}}) > \Pl(\intension{\phi_n}) = \bot$,
but this contradicts the assumption that the sequence
$\phi_1,\ldots,\phi_n$ is a counterexample to richness.
Since $\Pl(\intension{\phi_{n-1}}) = \bot$ in every structure that
satisfies $\Delta$, we conclude that $\Delta \sat_\PLclass \phi_{n-1} \Cond
\false$.

We now show that
$\Delta \not\vdashp \phi_{n-1} \Cond \False$. The easiest way of proving
this is by using \Tref{thm:pref-complete}. All we need to show is that
there is a preferential structure that satisfies $\Delta$ but does not
satisfies $\phi_{n-1} \Cond\False$. Let $W = \{ w_1,
\ldots, w_{n-1}\}$, let $\prec$ be such that $w_i \prec w_j$ for all $i <
j$, and let $\pi$ be such that $\pi(w_i)(\phi_{j}) =$ {\bf true} if and
only if $i = j$. It is straightforward to verify that
$(W,\prec,\pi)$ satisfies $\Delta$. However, it easy to see that
$(W,\prec,\pi) \not\sat \phi_{n-1} \Cond \false$. This concludes the
proof of ``if'' direction.

For the ``only if'' direction, assume that there is some
$\Delta$ and $\phi \Cond \psi$ such that $\Delta \sat_\PLclass
\phi\Cond\psi$ but $\Delta \not\vdashp \phi\Cond\psi$.
Using \Tref{thm:pref-complete} we get that $\Delta \not\satp
\phi\Cond\psi$. Thus, there is some preferential structure $P = (W,
\prec, \pi)$ that satisfies the defaults in $\Delta$ but not
$\phi\Cond\psi$. In fact, as the following lemma shows, we can assume that
$P$ is a {\em linear\/} structure.

\lem{\rm \cite{FrH3}}
\label{lem:linear}
Let $\Delta$ be a finite set of defaults.
If there is a preferential structure that satisfies $\Delta$ and does
    not satisfy $\phi\Cond\psi$,  then there is a preferential
structure $P = (W, \prec, \pi)$ such that $W = \{ w_1, \ldots, w_n
\}$, $w_i \prec w_j$ for all $i < j$, $P \sat \Delta$ and $P
\not\sat \phi\Cond\psi$.
\elem

We now use $P$ to construct a sequence of formulas that will be a
counterexample to the richness of $\PLclass$. Let $p_1,\ldots, p_m$ be
the propositions that appear in $\Delta$ and $\phi\Cond\psi$.
    Since $\Delta$ is finite, there is a finite number of such
propositions.
We note that whether a default
$\phi\Cond\psi$ is satisfied in $P$ depends only on the minimal world
satisfying $\phi$. If $\pi(w_i)$ and $\pi(w_j)$ for some $i < j$ agree
on the truth of
$p_1,\ldots,p_m$, then $w_j$ cannot be a minimal world for any formula
defined using only $p_1\ldots p_m$.
Thus, we can assume, without loss of generality, that for all $w_i \neq w_j$,
there is some $p_k$ that is assigned a different truth value by each
    of the two worlds.
We now construct formulas that characterize the truth assignment to
$p_1,\ldots, p_m$ in each world in $W$.
Let
$$\phi_i = \band_{
    \{j : \pi(w_i)(p_j) = {\bf true}\}}p_j \land \band_{ \{j :
\pi(w_i)(p_j) = {\bf false}\}}\neg p_j$$
for $i = 1, \ldots, n$,
and let
$\phi_{n+1} = \neg (\phi_1 \lor \ldots \lor \phi_n)$.
It is easy to verify that these formulas are mutually exclusive.

We now claim that if $\PL$ is a plausibility structure where
$\Pl(\intension{\phi_1}) > \cdots > \Pl(\intension{\phi_{n+1}}) =
\bot$, then $\PL$ satisfies the defaults in $\Delta$ but not $\phi
\Cond \psi$. This will suffice to prove that $\PLclass$ is not rich, since
if $\PLclass$ contains such a structure we get a contradiction to the
assumption that $\Delta \sat_\PLclass \phi\Cond\psi$.

Let $\PL$ be a plausibility structure where
$\Pl(\intension{\phi_1}) > \cdots > \Pl(\intension{\phi_{n+1}}) =
\bot$.
We want to show that $\PL \sat \xi\Cond\xi'$ if and only if $P \sat
\xi\Cond\xi'$, for all formulas $\xi$ and $\xi'$ defined using only
$p_1, \ldots, p_m$.

Let $\xi, \xi'$ be formulas defined over $p_1,\ldots, p_m$.
Assume that $P \sat \xi\Cond\xi'$. There are two cases: either (a) $\xi$
is not satisfied in $W$, or (b) the minimal world satisfying $\xi$ also
satisfies $\xi'$. In case (a),
it is easy to see that $\vdashL \xi \rimp \phi_{n+1}$. From A1, we
have that $\Pl(\xi) = \bot$, and thus $\PL \sat \xi\Cond \xi'$ vacuously.
In case (b) assume that $w_i$ is the minimal world satisfying $\xi$.
Since $P \sat
\xi \Cond \xi'$ we have that $\pi(w_i) \sat \xi \land \xi'$.
A simple argument shows that $\vdashL \phi_i \rimp
\xi\land\xi'$. Thus, using A1, we get that
$\Pl(\intension{\xi\land\xi'}) \ge \Pl(\intension{\phi_i})$.
Since $w_i$ is the minimal world satisfying $\xi$ and it also
satisfies $\xi'$, we have that $\xi\land\neg\xi'$ is not satisfied by
$w_1,\ldots, w_i$. Since $\phi_1,\ldots, \phi_{n+1}$ are exhaustive,
    we have that $\vdashL (\xi\land\neg\xi') \rimp
(\phi_{i+1} \lor \ldots \lor \phi_{n+1})$. Thus,
$\Pl(\intension{\xi\land\neg\xi'}) \le
\Pl(\intension{\phi_{i+1} \lor \ldots \lor \phi_{n+1}})$. By repeated
    applications of A2 and
the fact that $\Pl(\intension{\phi_i}) > \Pl({\intension{\phi_j}})$ for
all $j > i$,  we get that $\Pl(\intension{\phi_i}) >
\Pl{\intension{\xi\land\neg\xi'}})$. We conclude that
$\Pl(\intension{\xi\land\xi'}) > \Pl(\intension{\xi\land\neg\xi'})$
and thus $\PL \sat \xi\Cond\xi'$.

Now
assume that $P \not\sat \xi\Cond\xi'$. Thus, there is a minimal world
$w_i$ that satisfies $\xi$; moreover, $w_i$ does not satisfy $\xi'$.
This implies that $P \sat \xi\Cond\neg\xi'$ and $\Pl(\intension{\xi})
> \bot$. Applying the
proof in the previous paragraph, we have that
$\Pl(\intension{\xi\land\neg\xi'}) > \Pl(\intension{\xi\land\xi'})$,
and thus $\PL \not\sat \xi\Cond\xi'$.
\eprf

\subsection{Proofs for Section~\protect\ref{sec:epsilon}}
\label{prf:epsilon}

We now prove \Tref{thm:epsilon-equivalent}. We start
with two preliminary lemmas.

\lem\label{lem:epsilon}
Let $(W,\F,\Pl)$ be a
normal
qualitative plausibility space such that
$\F$ is finite, let $A^*, B^* \in \F$ be disjoint sets
such that $\Pl(A^*) \not< \Pl(B^*)$, and let $x \ge 2$. Then there is a
probability measure $\Pr$ over $W$ such that $\Pr(B^* | A^* \union B^*) \le
\frac{1}{2}$; moreover if $\Pl(A) < \Pl(B)$ then $\Pr(B | A \union B ) \ge
1 - \frac{1}{x}$, for all disjoint sets $A, B \in \F$.
\elem

\prf
Let $A_1, \ldots, A_n$ be the {\em atoms\/} of $\F$, \ie each $A_i \ne
\emptyset$ is in $\F$ and there is no nonempty $B \in \F$ such that
$B \subset A_i$. Since $\F$ is finite, every set in $\F$ is a disjoint
union of atoms.

We can describe $\Pl$ using a set of defaults.
Let $p_1, \ldots, p_n$ be a collection of distinct propositions.
For each set $A \in \F$
we define $\phi_A = \bor_{A_i \subseteq A} p_i$ if $A$ is nonempty,
and define $\phi_A = \False$ if $A$ is empty. Let
\begin{eqnarray*}
\Delta &= & \{ (\phi_A \lor \phi_B) \Cond \phi_B : A, B \in \F, A \inter
B = \emptyset, \Pl(A) < \Pl(B) \} \union \\
&& \{ (p_i \land p_j) \Cond\False : i \neq j \} \union \\
&& \{ \neg(p_1\lor\ldots\lor p_n) \Cond \False \}.
\end{eqnarray*}

Let $\pi$ be a truth assignment to $W$ such that $\pi(w)(p_i) = $ {\bf
true} if and only if $w \in A_i$. Then it is easy to check that
$\PL = (W,\F,\Pl,\pi)$ satisfies $\Delta$ and $\PL \not\sat
(\phi_{A^*} \lor \phi_{B^*}) \Cond \phi_{B^*}$.
Since $\Pl$ is a qualitative plausibility measure by assumption, we have
that $\PL \in \PLclass^{QPL}$.  Using \Tref{thm:main}, we get that
$\Delta \not\vdashp (\phi_{A^*} \lor \phi_{B^*}) \Cond \phi_{B^*}$.
By \Tref{thm:pref-complete}, there is a 
preferential
structure $P = (W_P, \prec, \pi_P)$ such that $P$ satisfies $\Delta$
but $P \not\sat
(\phi_{A^*} \lor \phi_{B^*}) \Cond \phi_{B^*}$. Applying
\Lref{lem:linear} we can assume that $P$ is linear, \ie $W_P =
\{v_1,\ldots, v_m\}$ for some $m \le n$, and $\prec$ is such that $v_i
\prec v_j$ if $i < j$.
Moreover, we have that $W_P$ not empty: Since $\PL$
is normal, we have that $\PL \not\sat \True \Cond \False$. Hence,
$P \not\sat \True \Cond \False$, which implies that
there are some minimal worlds that satisfy $\True$.

We now construct the required probability measure over $W$.
We start by noting that the defaults in $\Delta$
imply that for each world $v_i$, $\pi_P(v_i)$ assigns {\bf true} to
exactly one proposition.
Thus, without loss of generality, we can assume that $\pi_P(v_i)(p_j) = ${\bf
true} if and only if $i = j$.
We define $\Pr$ by assigning a probability to each atom. The
probability of all other sets is induced from this assignment:
$\Pr(A) = \sum_{A_i \subseteq A}\Pr(A_i)$, for all $A \in \F$.
Let $A_i$ be an atom. We define $\Pr(A_i) = \alpha \cdot
(\frac{1}{x})^i$, if $i \le m$; otherwise we define $\Pr(A_i) = 0$. The
constant $\alpha$ is a normalization constant that ensures that the
probability of atoms sum to 1. It is easy to verify that $\alpha =
(x-1)\frac{x^m}{x^m - 1}$.

We now  show that $\Pr$ satisfies the requirements of the
lemma. Assume that $A, B \in \F$ are disjoint and $\Pl(A) < \Pl(B)$.
We want to show that $\Pr(B | A \union B) \ge 1-\frac{1}{x}$.
By definition, $(\phi_A \lor \phi_B) \Cond \phi_B \in \Delta$, and thus
$P \sat (\phi_A \lor \phi_B) \Cond \phi_B$. There are two cases:
either (a) there is no world in $W_P$ that satisfies $(\phi_A \lor
\phi_B)$ or (b) the minimal world that satisfies $(\phi_A \lor
\phi_B)$ also satisfies $\phi_B$. In case (a), it immediately follows
that if $A_i \subseteq A \union B$, then $i > m$. We conclude
that $\Pr(A) = \Pr(B) = 0$. Thus, $\Pr(B|A \union B) = 1$. (Recall that
if $\Pr(A \union B) = 0$ then, by convention, $\Pr(B|A \union B) =
1$.)
In case (b), assume that $v_i$ is the minimal world
satisfying
$\phi_A \lor \phi_B$. Since $P \sat (\phi_A \lor \phi_B) \Cond \phi_B$,
it must be the case that $v_i$ satisfies $\phi_B$. This implies that
$\Pr(B) \ge \Pr(A_i)$. Since $v_i$ is the minimal world that satisfies
$\phi_A \lor \phi_B$, and since $v_i$ does not satisfy $\phi_B$
(since $A$ and $B$ are disjoint)
we conclude that $\Pr(A) \le \sum_{j > i} \Pr(A_j)$.
Simple calculation show  that
$\sum_{j > i} \Pr(A_j) \le \Pr(A_i)\cdot\frac{1}{x-1}$.
Thus, $(x - 1)\Pr(A) \le \Pr(A_i) \le \Pr(B)$. This implies that
$(x-1)(\Pr(A) + \Pr(B)) \le x\Pr(B)$. Since $A$ and $B$ are disjoint,
we have that $\Pr(A\union B) = \Pr(A) + \Pr(B)$. We get that
$(x-1)\Pr(A \union B) \le x \Pr(B)$, so
$\Pr(B | A \union B) \ge \frac{x-1}{x}$.

Finally we have to show that $\Pr(B^* | A^* \union B^*) \le \frac{1}{2}$.
Since $P \not\sat (\phi_{A^*} \lor \phi_{B^*}) \Cond \phi_{B^*}$, the
minimal world, $v_i$,  satisfying
$\phi_{A^*} \lor \phi_{B^*}$ does not satisfy $\phi_{B^*}$.
Since $A^*$ and $B^*$ are disjoint, this implies that $v_i$ satisfies
$\phi_{A^*}$. The argument above shows that $\Pr(A^* | A^*
\union B^*) \ge 1 - \frac{1}{x}$ and $\Pr(A^*\union B^*) > 0$.
Therefore, $\Pr(B^* | A^* \union B^*) \le \frac{1}{x} \le \frac{1}{2}$.
\eprf

\lem\label{lem:epsilon-equivalent}
Let $(W,\F,\Pl)$ be a
normal qualitative
plausibility space such that $\F$ is finite.
Then there is a PPD $PP = \{ \Pr_n : n \ge 1\}$ such that,
for all disjoint $A, B \in \F$,
\begin{itemize}\denselist
 \item if $\Pl(A) < \Pl(B)$, then $\Pr_n(B | A \union B) \ge 1 -
\frac{1}{n+1}$ for all $n$,
 \item if $\Pl(A) \not< \Pl(B)$, then for all $n$, there is an $m$
such that $n \le
    m < n + |\F|^2$ and $\Pr_m(B | A \union B) \le \frac{1}{2}$.
\end{itemize}
\elem
\prf
Let $A_0, \ldots, A_{|\F|-1}$ be some enumeration of the members of
    $\F$.
We define the PPD $PP = \{ \Pr_n : n \ge 1\}$ as follows. For each $n
\ge 1$ we define  $i_n,j_n$ to be the unique integers such that $n =
k\cdot|\F|^2 + i_n\cdot |\F| + j_n$ for some positive integer $k$.
According to \Lref{lem:epsilon}
there exists a probability distribution $\Pr_n$ on $W$ such that
if $\Pl(A) < \Pl(B)$, then $\Pr_n(B | A \union B) \ge 1 -
\frac{1}{n+1}$, for all disjoint sets $A, B \in F$.
Moreover, if $\Pl(A_{i_n}) \not< \Pl(A_{j_n})$, we can ensure that
$\Pr_n(A_{j_n} | A_{i_n} \union A_{j_n}) \le \frac{1}{2}$.
It is easy to verify that this PPD satisfies the requirements of
lemma.
\eprf

\rethm{thm:epsilon-equivalent}
If $\PL \in \PLclass^{QPL}$ is a
normal plausibility structure
for a countable language $\L$,
then there is a structure $\PL' \in
\PLclass^{\epsilon}$ 
that is
default-equivalent to $\PL$.
\erethm
\prf
Let $\PL = (W, \F, \Pl, \pi)$ be a normal qualitative plausibility
for a countable language $\L$.  Since $\L$ is countable, so is $\F=
\{\intension{\phi}: \phi \in\L\}$.  Let $A_1, A_2, \ldots$ be an
enumeration of the sets in
$\F$.
Without loss of generality, we can assume that $A_1 = W$.  Since
$\Pl$ is normal, we must have $\Pl(A_1) > \bot$.
For each
$k$, let $\F_k$ be the minimal algebra that contains $A_1, \ldots,
A_k$. Clearly, $\F_k$ is a finite algebra,
and $\Pl$ restricted to $\F_k$ is normal.
Thus, there is a PPD
$PP^k = \{\Pr^k_1, \Pr^k_2, \ldots \}$ that satisfies the conditions of
\Lref{lem:epsilon-equivalent}.

We now use elements of these sequences to construct the desired PPD.
We define $PP$ to be the sequence that consists of a
segment of $PP^1$, followed by a segment of $PP^2$, and so on, such
that the length of the segment from $PP^k$ is $|F_k|^2$:
$$\{ {\Pr}^1_{1}, \ldots, {\Pr}^1_{1 + |\F_1|^2 -
1}, {\Pr}^2_{2}, \ldots, {\Pr}^2_{2 + |\F_2|^2-1}, \ldots,
{\Pr}^k_{k}, \ldots, {\Pr}^k_{k + |\F_k|^2-1}, \ldots \}.$$
To show that the plausibility measure that corresponds to $(W, PP,
\pi)$ is default-equivalent to $\PL$ we need to show that for all
disjoint $A_i,A_j \in \F$, $\Pl(A_i)
    < \Pl(A_j)$ if and only if $\lim_{n\rightarrow\infty} \Pr_n(A_j |
    A_i \union A_j) = 1$.

Assume that $\Pl(A_i) < \Pl(A_j)$. Let $\epsilon > 0$. Set $m_\epsilon
= \max(i,j, \lceil \frac{1}{\epsilon} \rceil)$, and $N_\epsilon =
\sum_{k = 1}^{m_\epsilon} |\F_k|^2$. Let $n > N_\epsilon$. Then our
construction is such that $\Pr_n = \Pr^k_l$ for some $l \ge k \ge
m_\epsilon$. Since $k \ge max(i,j)$, we have
that $A_i, A_j \in \F_k$. Moreover, according to
\Lref{lem:epsilon-equivalent}, $\Pr_{l}^k(A_j | A_i \union A_j) \ge
1 - \frac{1}{l+1} \ge 1 - \frac{1}{m_\epsilon+1} \ge 1 - \epsilon$.
Thus, we conclude that $\lim_{n\rightarrow\infty} \Pr_n(A_j | A_i
\union A_j) = 1$.

Assume that $\Pl(A_i) \not< \Pl(A_j)$. Let $k > \max(i,j)$.
According to \Lref{lem:epsilon-equivalent}, there is an $m$ such that
$k \le m < k + |\F_k|^2$ such that $\Pr^k_m(A_j|A_i \union A_j) \le
\frac{1}{2}$. Moreover, by the our construction, there is an $n$ such
that $\Pr_n = \Pr^k_m$. Thus, for infinitely many $n$,
$\Pr_n(A_j|A_i \union A_j) \le \frac{1}{2}$.
We conclude that $\lim_{n\rightarrow\infty} \Pr_n(A_j | A_i
\union A_j) \neq 1$.
\eprf

\subsection{Proofs for Section~\protect\ref{entrenchment}}
\label{prf:entrenchment}

\repro{pro:epistemic}
If $E$ is an expectation structure, then $\PL_E$ is a plausibility
structure. Furthermore, $E \sat \phi\Cond\psi$ if
    and only if $\PL_E \sat \phi\Cond\psi$.
\erepro

\prf
Suppose $E = (\L,\Ele)$ is an expectation structure, and $\PL_E =
(W_E,\Pl_E,\pi_E)$.
To show that $\PL_E$ is a plausibility structure, we need to check
that the ordering defined by $\Pl_E$ is reflexive, transitive, and
satisfies A1. Note that it is easy to show, using standard arguments,
that $\intension{\phi} \subseteq \intension{\psi}$ if and only if
$\vdashL \phi \rimp \psi$. The proof that $\PL_E$ is plausibility
measure follows in a straightforward manner and is left as an exercise
to the reader.

We now show that $E \sat \phi\Cond\psi$ if and only if $\PL_E
\sat \phi\Cond\psi$.
Assume that $E \sat \phi\Cond\psi$. According to \Tref{thm:MakGar} there
are two possible cases: either (a) $\vdashL \phi \rimp\psi$ or (b)
$(\phi\rimp \neg\psi) \Elt (\phi\rimp\psi)$.
In case (a), we apply REF and RW (which
are valid in all plausibility structures) to get that $\PL_E
\sat \phi\Cond\psi$.
Now consider case (b). It is clear
that $\True \not\!\!\Ele (\phi\rimp\neg\psi)$, for if $\True \Ele
(\phi\rimp\neg\psi)$, then $\True \Elt
(\phi\rimp\psi)$ by E1. We also have $(\phi\rimp\psi) \Ele \True$ by
using E2, so by E1 again, we get that $\True \Elt \True$, which is a
contradiction.
Since $\True \not\!\!\Ele (\phi\rimp\neg\psi)$ and $(\phi\rimp\neg\psi)
\Elt (\phi\rimp\psi)$, by definition, we have that
$\Pl_E(\intension{\phi\land\neg\psi}) \le
\Pl_E(\intension{\phi\land\psi})$. Moreover, since
$\True \not\!\!\Ele (\phi\rimp\neg\psi)$, we have that $\phi\land\psi$ is
consistent. Thus, $\not\vdashL \phi\land\psi \rimp \phi\land\neg\psi$.
Furthermore, since $(\phi\rimp \neg\psi) \Elt (\phi\rimp\psi)$, we
also have that $(\phi\rimp \psi) \not\!\!\Elt (\phi\rimp\neg\psi)$. This
implies, by definition, that $\Pl_E(\intension{\phi\land\neg\psi}) \not\ge
\Pl_E(\intension{\phi\land\psi})$.
Hence, $\Pl_E(\intension{\phi\land\neg\psi}) <
\Pl_E(\intension{\phi\land\psi})$, and so $\PL_E \sat \phi\Cond\psi$.

Now assume that $\PL_E \sat \phi\Cond\psi$. If $\vdashL \phi\rimp\psi$,
then $E \sat \phi\Cond\psi$. Assume that $\not\vdashL \phi\rimp\psi$.
This implies that $\phi\land\neg\psi$ is consistent in $\L$. We claim that
$\Pl_E(\intension{\phi}) > \bot$. To see this, assume that
$\Pl_E(\intension{\phi} \le \Pl(\intension{\False})$. Since $\True
\Ele \neg\False$, the definition of $\Pl_E$ implies that $\vdashL \phi \rimp
\False$. But this contradict the assumption that $\phi$ is consistent
in $\L$. We conclude that $\Pl_E(\intension{\phi}) > \bot$, and thus
since $\PL_e \sat \phi\Cond\psi$,
$\Pl_E(\intension{\phi\land\psi}) >
\Pl_E(\intension{\phi\land\neg\psi})$. It is straightforward to show
that since $\phi\land\neg\psi$ is consistent in $\L$, we get that
$\neg(\phi\land\psi) \Elt \neg(\phi\land\neg\psi)$. Rewriting this
equation, we get that $(\phi\rimp\neg\psi) \Elt (\phi\rimp\psi)$, and
thus $E \sat \phi\Cond\psi$.
\eprf

\subsection{Proofs for Section~\protect\ref{conditionallogic}}
\label{prf:conditionallogic}

We now want to prove Theorem~\ref{SysCcomplete}.  For the proof, it is
useful to define  $\PBox \phi$ as an abbreviation for
$\neg\phi\Cond\False$.
(The $\PBox$ operator is called the {\em outer modality\/} in
\cite{Lewis73}.)  Expanding the definition of $\Cond$, we get that
$\PBox\phi$ holds at $w$ if and only if $\Pl(\intension{\neg\phi}) =
\bottom$. Thus, $\PBox\phi$ holds if $\neg\phi$ is considered
completely implausible. Thus, it implies that $\phi$ is true ``almost
everywhere''.  The following lemma collects some properties of
$\PBox$ that will be needed in the proof.

\lem\label{lem:delta}
\begin{itemize}\denselist
 \item[(a)] $\vdashc \PBox(\phi\land\psi) \rimp \PBox\phi$.
\item[(b)] $\vdashc (\PBox\phi \land \PBox\psi) \rimp
\PBox(\phi \land \psi)$.
 \item[(c)] $\vdashc (\PBox\phi\land \PBox(\phi\rimp\psi)) \rimp \PBox\psi$.
 \item[(d)] If $\vdashc \phi$ then $\vdashc \PBox\phi$.
 \item[(e)] $\vdashc \PBox\phi \rimp (\psi\Cond\phi)$.
 \item[(f)] $\vdashc (\PBox(\phi \dimp \phi') \land \PBox(\psi \dimp
\psi')) \rimp ((\phi \Cond\psi) \dimp (\phi'\Cond\psi'))$.
\end{itemize}
\elem
\prf
Recall that $\PBox\phi$ is defined as $\neg\phi\Cond \False$.

We start with part (a).
\begin{enumerate}\denselist

 \item[1.] $\vdashc (\neg(\phi\land\psi)\Cond\False) \rimp
(\neg(\phi\land\psi) \Cond \neg\phi)$ \hfill RC2
 \item[2.] $\vdashc (\neg(\phi\land\psi)\Cond\False) \rimp
((\neg(\phi\land\psi) \land \neg\phi) \Cond \false)$ \hfill 1, C4
 \item[3.]  $\vdashc (\neg(\phi\land\psi)\Cond\False) \rimp
(\neg\phi \Cond \false)$ \hfill 2, RC1
 \item[4.] $\vdashc \PBox(\phi\land\psi) \rimp \PBox\phi$ \hfill 3
rewritten
\end{enumerate}

To prove part (b), we proceed as follows:
\begin{enumerate}\denselist
\item[1.] $\vdashc ((\neg\phi\Cond\False)\land (\neg\psi
\Cond\False)) \rimp ((\neg\phi\lor\neg\psi) \Cond\False)$
\hfill C3
\item[2.] $\vdashc ((\neg\phi\Cond\False)\land (\neg\psi
\Cond\False)) \rimp (\neg(\phi\land\psi) \Cond\False)$ \hfill 1, RC1
\item[3.] $\vdashc (\PBox\phi \land\PBox \psi) \rimp
\PBox(\phi\land\psi)$ \hfill 2 rewritten
\end{enumerate}

For part (c), we proceed as follows:
\begin{enumerate}\denselist
\item[1.] $\vdashc (\PBox \phi \land \PBox(\phi \rimp \psi)) \rimp
\PBox(\phi \land (\phi \rimp \psi))$ \hfill (b)
\item[2.] $\vdashc (\PBox \phi \land \PBox(\phi \rimp \psi)) \rimp
\PBox(\phi \land \psi)$ \hfill 1, RC1
\item[3.] $\vdashc (\PBox \phi \land \PBox(\phi \rimp \psi)) \rimp
\PBox\psi$ \hfill (a), 2
\end{enumerate}

To prove part (d), we assume that $\vdashc \phi$.
\begin{enumerate}\denselist

 \item[1.] $\vdashc \phi$ \hfill assumption

 \item[2.] $\vdashc \true \dimp \phi$ \hfill 1, C0
 \item[3.] $\vdashc \False \Cond\False$ \hfill C1
 \item[4.] $\vdashc \neg\phi \Cond\False$ \hfill 3, RC1
\end{enumerate}

To prove part (e), we proceed as follows:
\begin{enumerate}\denselist
 \item[1.] $\vdashc (\psi\land\phi) \Cond \phi$ \hfill C1, RC2
 \item[2.] $\vdashc (\neg \phi \Cond \False) \rimp
(\neg\phi \Cond \phi) \land (\neg \phi \Cond \psi)$ \hfill RC2
\item[3.] $((\neg\phi \Cond \phi) \land (\neg \phi \Cond \psi)) \rimp
((\psi \land \neg \phi) \Cond \phi$ \hfill C4
\item[4.] $\vdashc (((\psi\land\phi) \Cond \phi)
\land ((\psi \land \neg \phi) \Cond \phi))) \rimp (\psi \Cond \phi)$
\hfill C3, RC1
 \item[5.] $\vdashc \PBox\phi \rimp (\psi \Cond\phi)$ \hfill 1, 2, 3, 4
\end{enumerate}

Finally, we prove part (f).
\begin{enumerate}\denselist
 \item[1.] $\vdashc \PBox(\phi\dimp\phi') \rimp (\PBox(\phi' \rimp \phi)
\land \PBox(\phi \rimp \phi'))$
\hfill (a)
\item[2.] $\vdashc \PBox(\phi' \rimp \phi) \rimp ((\phi' \land \neg \phi)
\Cond \psi)$ \hfill definition of $\PBox$, RC2
 \item[3.] $\vdashc \PBox(\phi\rimp\phi') \rimp (\phi \Cond (\phi
\rimp \phi'))$ \hfill (e)
 \item[4.] $\vdashc \PBox(\phi\rimp\phi') \rimp (\phi \Cond \phi')$
\hfill 3, C1, C2, RC2
 \item[5.] $\vdashc (\phi\Cond\phi')\land(\phi\Cond\psi) \rimp
((\phi'\land\phi) \Cond \psi)$ \hfill C4
 \item[6.] $\vdashc \PBox(\phi\dimp\phi')\land(\phi\Cond\psi) \rimp
(\phi'\Cond\psi)$ \hfill 1, 2, 3, 4, 5, C3, RC1

 \item[7.] $\vdashc\PBox(\psi\dimp\psi') \rimp (\phi' \Cond
(\psi\rimp\psi'))$ \hfill (e), RC2
 \item[8.] $\vdashc \PBox(\phi\dimp\phi')\land\PBox(\psi\dimp\psi')
\land (\phi\Cond\psi) \rimp (\phi'\Cond\psi')$ \hfill 6, 7, C2, RC2
\end{enumerate}
\eprf

\rethm{SysCcomplete}
System {\bf C} is a sound and complete axiomatization of $\LCond$
with respect to
$\PLclass^{QPL}_c$.
\erethm
\prf
It is easy to verify that System~{\bf C} is sound in $\PLclass_c^{QPL}$.
To prove completeness, we have to show that if $\sat_{\PLclass_c^{\sQPL}}
\phi$, then $\vdashc \phi$. This is equivalent to showing that if
$\not\vdashc \phi$ (\ie $\neg\phi$ is consistent) then
$\not\sat_{\PLclass_c^{\sQPL}} \phi$ (\ie $\neg\phi$ is satisfiable).

We construct a {\em canonical\/} qualitative plausibility
structure $\PL$ such that for all $\xi \in \LCond$ we have
that if $\xi$ is consistent, then $(\PL,w) \sat \xi$ for some world $w$,
using standard techniques.
Recall that a set of formulas $V \subseteq
\LCond$ is a {\em maximal $\vdashc$-consistent\/} set if it is
consistent with respect to $\vdashc$ and for each $\phi \in \LCond$, either $\phi
\in V$ or $\neg \phi \in V$. Let $V$ be a set of formulas. We define
    $V/\PBox = \{ \phi : \PBox\phi \in V \}$.

Define $\PL = (W, \Plass, \pi)$ as follows:
\begin{itemize}\denselist
 \item
$W = \{ w_V : V \subseteq \LCond$ is a maximal $\vdashc$-consistent
set of formulas $\}$,
 \item $\Plass(w_V) = (W_w,\F,\Pl_{w_V})$ where
\begin{itemize}\denselist
 \item $\Omega_{w_V} = \{ w_U : V/\PBox \subseteq U \}$,
  \item $\F_{w_V} = \{ \IntensionV{\phi} :  \phi \in
    \LCond\}$, where  $\IntensionV{\phi} = \{ w_{U} \in W_{w_V} :
\phi \in U \}$,
  \item $\Pl_{w_V}$ is such that $\Pl_{w_V}(\IntensionV{\phi}) \le
\Pl_{w_V}(\IntensionV{\psi})$ if and only if $(\phi\lor\psi) \Cond \psi \in
V$,
\end{itemize}
  \item $\pi(w_V)(p)  = $ {\bf true} if and only if $p \in V$.
\end{itemize}
We want to show that $\PL$ is a qualitative plausibility structure
and that
$(\PL,w_V) \sat \phi$ if and only if $\phi \in V$ for
all formulas $\phi$ and worlds $w_V$.

We first need to establish that $\Pl_{w_V}$ is well-defined, for all
$w_V \in W$.
Let $V$ be a maximal $\vdashc$-consistent set.
We need to show that if $\IntensionV{\phi} = \IntensionV{\phi'}$ and
$\IntensionV{\psi} = \IntensionV{\psi'}$, then $(\phi\lor\psi) \Cond \psi
    \in V$ if and only $(\phi'\lor\psi') \Cond \psi' \in V$.

We claim that $\IntensionV{\phi} = \IntensionV{\phi'}$ if and only if
$V/\PBox \union \{ \neg(\phi \dimp \phi') \}$ is inconsistent.
The ``if'' direction is obvious.  For the ``only if'' direction, assume
that $V/\PBox \union \{ \neg(\phi \dimp \phi') \}$ is consistent.
This implies that there is a maximal consistent set $U$ such that
$\Delta \union \{ \neg(\phi \dimp \phi') \} \subseteq U$.
Clearly, $w_U$ is in the symmetric difference between $\IntensionV{\phi}$
and $\IntensionV{\phi'}$. Thus, $\IntensionV{\phi} \neq
\IntensionV{\phi'}$.

Now assume that $\IntensionV{\phi} = \IntensionV{\phi'}$ and
$\IntensionV{\psi} = \IntensionV{\psi'}$. Therefore, as we have just
shown, $V/\PBox \union \{
\neg(\phi \dimp \phi') \}$ and $V/\PBox \union \{ \neg(\psi \dimp \psi)
\}$ are both inconsistent. Thus, there exists a formula $\delta$ which
is the conjunction of a finite number of formulas in $V/\PBox$
such that $\vdashc
\delta \rimp (\phi \dimp \phi') \land (\psi
\dimp \psi)$.
Using part (b) of Lemma~\ref{lem:delta}, and the fact that if $\phi
\in V/\PBox$, then $\PBox\phi \in V$, we get that $\PBox \delta \in
V$. From parts
(c) and (d) of Lemma~\ref{lem:delta}, we get that
$\vdashc \PBox\delta \rimp \PBox(\phi \dimp \phi') \land \PBox(
\psi \dimp \psi')$.
Finally, using part (f) of \Lref{lem:delta} we get that
$$
\vdashc \PBox\delta \rimp ((\phi\Cond\psi) \dimp (\phi'\Cond\psi')).$$
Thus, $\phi\Cond\psi \in V$ if and only if
$\phi'\Cond\psi' \in V$. This suffices to prove that $\Pl_{w_V}$ is
well-defined.

To see that $\PL$ is a qualitative
plausibility structure, first note that the definition of $\Pl_{w_V}$
mirrors the
construction in \Lref{lem:embedding}. It is easy to prove that the set
of defaults
$\{ \phi \Cond \psi \in V \}$ is closed under the KLM rules.
Thus, we can immediately use the proof of \Lref{lem:embedding} to
show that $(W_{w_V},\Pl_{w_V})$ is a plausibility space.
The proof that $\PL$ satisfies A2 and A3 is
identical to the proof of \Tref{QPL}.  Thus,
$\PL \in \PLclass_c^{QPL}$.

We next  show that $(\PL,w_V) \sat \phi$ if and only if $\phi
    \in V$. This is done by induction on the structure of $\phi$. The
    only case of interest is if $\phi$ is of the form $\phi'\Cond\psi$.
    Here again the proof is identical to the proof of
\Lref{lem:embedding}. The only difference is the use of axioms in
\SysC\ instead of the corresponding rules in \SysP.

\commentout{
Assume
    $\phi\Cond\psi \in V$. Then, using C1, LLE, and RW we get that
    $(\phi\land\psi) \lor (\phi\land\neg\psi) \Cond (\phi\land\psi)
    \in V$. Thus, $\Pl( \IntensionV{\phi\land\psi} ) \ge
    \Pl( \IntensionV{\phi\land\neg\psi})$. Since
    $\phi\land\psi$ and $\phi\land\neg\psi$ are disjoint, we conclude
    that the two plausibilities are equal if and only if
    $\IntensionV{\phi} = \emptyset$. Using the induction
    hypothesis we conclude that either
    $\Pl(\intension{\phi}) = \bottom$ or
    $\Pl(\intension{\phi\land\psi}) >
    \Pl(\intension{\phi\land\neg\psi})$.
Assume $(\PL,w_V) \sat \phi\Cond\psi$. There are two cases, either
    $\Pl(\intension{\phi}) = \bottom$ or
    $\Pl(\intension{\phi\land\psi}) >
    \Pl(\intension{\phi\land\neg\psi})$. In the
    first case, we use the induction hypothesis to conclude that
    $\Pl(\IntensionV{\phi}) = \bottom$. This
    implies that $\phi\Cond\False \in V$. Using RW we conclude that
    $\phi\Cond\psi \in V$. In the second case, we use the induction
    hypothesis to conclude that $\Pl(
    \IntensionV{\phi\land\psi} ) > \Pl(
    \IntensionV{\phi\land\neg\psi})$. This implies that
    $(\phi\land\psi) \lor (\phi\land\neg\psi) \Cond (\phi\land\psi)
    \in V$. Using LLE and RW we conclude that $\phi\Cond\psi \in
    V$.
}

Let $\xi$ be a $\vdashc$-consistent formula.  Using standard
arguments, it is easy to show that there is some maximal
$\vdashc$-consistent set $V_\xi$ such that $\xi \in V_\xi$. Thus,
$(\PL, w_{V_\xi}) \sat
\xi$, so $\xi$ is satisfiable in $\PLclass_c^{QPL}$.
\eprf

\repro{pro:rational}
Let $\P \subseteq \PLclass_c^{QPL}$. C5 is valid in $\PLclass$ if and only if all
structures in $\PLclass$ are rational.
\erepro

\prf
To prove the ``if'' direction it suffices to show that C5 is sound in
rational qualitative structures. Let $\PL = (W,\Plass,\pi)$ be a
rational qualitative plausibility structure. Assume that
$(\PL,w) \sat (\phi\Cond \psi)$ and $(\PL,w) \sat \neg(\phi\land\xi
\Cond \psi)$. We need to prove that $(\PL,w) \sat \phi \Cond \neg\xi$.
If $\Pl_w(\intension{\phi}_{(\sPL,w)}) = \bot$, then $(\PL,w) \sat
\phi \Cond \neg \xi$ vacuously, and we are done. Now assume that
$\Pl_w(\intension{\phi}_{(\sPL,w)}) > \bot$. Let $A =
\intension{\phi}_{(\sPL,w)}$, $B = \intension{\psi}_{(\sPL,w)}$, and
$C = \intension{\xi}_{(\sPL,w)}$. Since $(\PL,w) \sat \phi\Cond \psi$,
we have that $\Pl_w(A \inter B) > \Pl_w(A \inter\overline{B})$. Since
$\Pl_w$ satisfies A5, we have that either $\Pl_w(A \inter B \inter C ) >
\Pl_w(A \inter\overline{B})$ or $\Pl_w(A \inter B \inter \overline{C} ) >
\Pl_w(A \inter\overline{B})$. However, since $(\PL,w) \sat
\neg(\phi\land\xi \Cond \psi)$, we have that $\Pl_w( A \inter B \inter
C ) \not> \Pl_w( A \inter \overline{B} \inter C )$. This implies that
\beqn
\Pl_w( A \inter B \inter C ) \not> \Pl_w( A \inter \overline{B} ).
\label{eq:rational-1}
\eeqn
Thus, we conclude that
\beqn
\Pl_w( A \inter B \inter \overline{C} ) > \Pl_w( A \inter \overline{B}
). \label{eq:rational-2}
\eeqn
Applying A4 with (\ref{eq:rational-2}) as the antecedent, we get
that either $\Pl_w( A \inter B \inter C ) > \Pl_w( A \inter
\overline{B} )$ or $\Pl_w(A \inter B \inter \overline{C}) > \Pl_w( A
\inter B \inter C )$. Since the former contradicts
(\ref{eq:rational-1}), we conclude that $\Pl_w(A \inter B \inter
\overline{C}) > \Pl_w( A \inter B \inter C )$. Using A1, A2, and
(\ref{eq:rational-2}), we get that $\Pl_w(A \inter \overline{C}) \ge
\Pl_w(A \inter B \inter  \overline{C}) > \Pl_w( (A \inter B \inter C)
\union (A \inter \overline{B}) ) \ge \Pl_w(A \inter C)$. Thus, $(\PL,w)
\sat \phi \Cond \neg \xi$.

To prove the ``only if'' direction, we have to show that if there is
some $\PL = (W,\Plass, \pi)$ in $\PLclass$ that is not rational, then C5 is
violated.

Suppose that $\PL$ does not satisfy A4. Since we have assumed that $\F_w
= \{ \intension{\phi}_{(\sPL,w)} : \phi\in\L \}$,  there is a world
$w$, and
formulas $\phi$, $\psi$, and
$\xi$ such that $\intension{\phi}_{(\sPL,w)}$,
$\intension{\psi}_{(\sPL,w)}$, and $\intension{\xi}_{(\sPL,w)}$ are
pairwise disjoint, $\Pl_w(\intension{\phi}_{(\sPL,w)}) >
\Pl_w(\intension{\psi}_{(\sPL,w)})$, and yet
$\Pl_w(\intension{\xi}_{(\sPL,w)}) \not >
\Pl_w(\intension{\psi}_{(\sPL,w)})$ and
$\Pl_w(\intension{\xi}_{(\sPL,w)}) \not<
\Pl_w(\intension{\phi}_{(\sPL,w)})$.
Since $\Pl_w(\intension{\phi}_{(\sPL,w)}) >
\Pl_w(\intension{\psi}_{(\sPL,w)})$, we have that
$\Pl_w(\intension{\phi \lor \xi}_{(\sPL,w)}) >
\Pl_w(\intension{\psi}_{(\sPL,w)})$. Thus, $(\PL,w) \sat (\phi \lor
\psi \lor \xi) \Cond \neg\psi$. Moreover, since
$\Pl_w(\intension{\xi}_{(\sPL,w)}) \not<
\Pl_w(\intension{\phi}_{(\sPL,w)})$, we have that
$\Pl_w(\intension{\xi}_{(\sPL,w)}) > \bot$. Since we also have that
$\Pl_w(\intension{\xi}_{(\sPL,w)}) \not >
\Pl_w(\intension{\psi}_{(\sPL,w)})$, we conclude that $(\PL,w) \sat
\neg(( \xi \lor \psi ) \Cond \neg\psi)$.
Finally, since $\Pl_w(\intension{\phi}_{(\sPL,w)}) \not>
\Pl_w(\intension{\xi}_{(\sPL,w)})$, we have that
$\Pl_w(\intension{\phi}_{(\sPL,w)}) \not>
\Pl_w(\intension{\xi\lor \psi}_{(\sPL,w)})$. Thus,
$(\PL,w) \sat \neg ((\phi \lor \psi \lor \xi) \Cond
\neg(\psi \lor \xi))$.
Define $\alpha$ as $\phi \lor \psi
\lor \xi$, $\beta$ as $\psi \lor \xi$ and $\gamma$ as $\neg \psi$. We
have
just proved that $(\PL,w) \sat (\alpha \Cond \gamma) \land \neg(
\alpha \Cond \neg\beta) \land \neg(\alpha
\land \beta \Cond \gamma)$, which
contradicts C5.

Now suppose that $\PL$ does not satisfy A5. Then there is a world
$w$, and
formulas $\phi$, $\psi$, and
$\xi$, such that $\intension{\phi}_{(\sPL,w)}$,
$\intension{\psi}_{(\sPL,w)}$, and $\intension{\xi}_{(\sPL,w)}$ are
pairwise disjoint, $\Pl_w(\intension{\phi\lor\psi}_{(\sPL,w)}) >
\Pl_w(\intension{\xi}_{(\sPL,w)})$, and yet
$\Pl_w(\intension{\phi}_{(\sPL,w)}) \not >
\Pl_w(\intension{\xi}_{(\sPL,w)})$ and
$\Pl_w(\intension{\psi}_{(\sPL,w)}) \not>
\Pl_w(\intension{\xi}_{(\sPL,w)})$.
Since $\Pl_w(\intension{\phi\lor\psi}_{(\sPL,w)}) >
\Pl_w(\intension{\xi}_{(\sPL,w)}) \ge \bot$, either
$\Pl_w(\intension{\phi}_{(\sPL,w)}) > \bot$ or
$\Pl_w(\intension{\psi}_{(\sPL,w)}) > \bot$, for otherwise,
using A3 we would get that $\Pl_w(\intension{\phi\lor\psi}_{(\sPL,w)})
= \bot$. Without loss of generality, we assume that
$\Pl_w(\intension{\phi}_{(\sPL,w)}) > \bot$.
Since  $\Pl_w(\intension{\phi\lor\psi}_{(\sPL,w)}) >
\Pl_w(\intension{\xi}_{(\sPL,w)})$, we have that $(\PL,w) \sat (\phi
\lor \psi \lor \xi) \Cond \neg \xi$.
Since $\Pl_w(\intension{\phi}_{(\sPL,w)}) > \bot$ and
$\Pl_w(\intension{\phi}_{(\sPL,w)}) \not >
\Pl_w(\intension{\xi}_{(\sPL,w)})$, we have that $(\PL,w) \sat \neg
((\phi \lor \xi) \Cond \neg\xi)$.
Finally, since $\Pl_w(\intension{\psi}_{(\sPL,w)})
\not> \Pl_w(\intension{\xi}_{(\sPL,w)})$, we have that
$\Pl_w(\intension{\psi}_{(\sPL,w)})
\not> \Pl_w(\intension{\phi\lor\xi}_{(\sPL,w)})$, and thus
$(\PL,w) \sat \neg((\phi \lor \psi \lor \xi) \Cond \psi))$.
 Define $\alpha$ as $\phi \lor \psi
\lor \xi$, $\beta$ as $\neg\psi$ and $\gamma$ as $\neg \xi$. We have
proved that $(\PL,w) \sat (\alpha \Cond \gamma) \land \neg( \alpha
\Cond \neg\beta) \neg(\alpha
\land \beta \Cond \gamma)$, which
contradicts C5.
\eprf

We next want to prove Theorem~\ref{thm:Ranked}.  We first need a lemma.

\lem\label{lem:<^*}
Let $(W,\Pl)$ be a qualitative plausibility space, and let $<^*$ be a binary
relation on subsets of $W$ such that $A <^* B$ if there
is some set $C \subseteq W$ such that $A \inter C =
\emptyset$ and $\Pl(A) < \Pl(C) \le \Pl(B)$. Then
\begin{itemize}\denselist
 \item[(a)] $<^*$ is a strict order; that is irreflexive, transitive and anti-symmetric,
 \item[(b)] if $A \inter B = \emptyset$, then $A <^* B$ if and only if
$\Pl(A) < \Pl(B)$,
 \item[(c)] if $\Pl$ is rational, then $<^*$ is modular, and

 \item[(d)] if $\Pl$ is rational and $A <^* (A \union B)$, then $B
 \not<^* (A \union B)$.
\end{itemize}
\elem

\prf
We start with part (a). The definition of $<^*$ implies that $A <^* B$ only
if $\Pl(A) < \Pl(B)$. Thus, we get that $<^*$ is irreflexive and
anti-symmetric. We now show that $<^*$ is transitive. Assume that $A
<^* B$ and $B <^* C$. Since $A <^* B$, there is a set $D$ such that $A
\inter D = \emptyset$, and $\Pl(A) < \Pl(D) \le \Pl(B)$. Moreover,
since $B <^* C$, we have that $\Pl(B) < \Pl(C)$. Thus, we get that
$\Pl(D) < \Pl(C)$, and hence $A <^* C$.

For part (b), let $A$ and $B$ be disjoint sets. If $\Pl(A) <
\Pl(B)$, then $A <^* B$ using the set $C = B$. On the other hand, if
$A <^* B$, then $\Pl(A) < \Pl(B)$.

For part (c), assume that $\Pl$ is rational, and that $A
<^* B$. Let $C \subseteq W$. We have to show that either $A <^* C$ or
$C <^* B$. Since $A <^* B$, there is a set $D$ such that $A \inter D =
\emptyset$ and $\Pl(A) < \Pl(D) \le \Pl(B)$. Since $D$ is the disjoint
union of $(D \inter C)$ and $D - C$, we can apply A5 and get that
either $\Pl(A) < \Pl(D \inter C)$ or $\Pl(A) < \Pl(D - C)$. If
$\Pl(A) < \Pl(D \inter C)$, then since $\Pl(D \inter C) \le \Pl(C)$,
we get that $A <^* C$ and we are done. If $\Pl(A) < \Pl(D - C)$, since $A$,
$D - C$ and $C - A$ are pairwise disjoint, we get from A4 that either
$\Pl(A) < \Pl(C - A)$ or $\Pl(C - A) < \Pl(D - C)$. If $\Pl(A) < \Pl(C
- A)$, we again get that $A <^* C$ and we are done. If $\Pl(C - A) < \Pl(D -
C)$, using A2, we get that $\Pl( (C - A) \union A ) < \Pl( D - C
)$. We conclude that $\Pl(C) < \Pl(D - C)$. Since $\Pl(D - C) \le
\Pl(D) \le \Pl(B)$, we get that $C <^* B$.

Finally, we prove part (d). Assume that $A <^* (A \union B)$.
Without loss of generality, we can also assume that $B \inter A =
\emptyset$ (if not, replace $B$ by $B - A$).
We want to show that $B \not<^* (A \union B)$. By way of
contradiction, assume that $B <^* (A \union B)$.
Since $A <^* B$, there is a set $C$ such that $A
\inter C = \emptyset$ and
$\Pl(A) < \Pl(C) \le \Pl(A\union B)$.
Since $C \subseteq (C - B) \union B$,
this implies that $\Pl(A) <
\Pl((C - B) \union B)$. Since $A$, $B$, and $C - B$ are pairwise
disjoint, we can apply A5 to get that either $\Pl(A) < \Pl(B)$ or
$\Pl(A) < \Pl( C - B)$. If $\Pl(A) <
\Pl(B)$, since we assumed that $B <^* (A\union B)$, there is a set $D$
such that $D \inter B = \emptyset$ and $\Pl(B) < \Pl(D) \le
\Pl(A\union B)$. This implies
that $\Pl(B) < \Pl( (D - A) \union B )$. Moreover, since $\Pl(A)
< \Pl(B)$, we also have that $\Pl(A) < \Pl( (D - A) \union B)$. Since
$A$, $B$ and $D - A$ are pairwise disjoint, we can apply A2 to get
that $\Pl(A \union B) < \Pl(D - A) \le \Pl(D)$. But this contradicts
the assumption that $\Pl(D) \le \Pl(A \union B)$. If $\Pl(A) < \Pl(C -
B)$, since $A$, $B$, and $C - B$ are pairwise disjoint, we have by A4
that either $\Pl(A) < \Pl(B)$ or $\Pl(B) < \Pl(C - B)$. We have
already seen that if $\Pl(A) < \Pl(B)$ we get a contradiction. Now
assume that $\Pl(B) < \Pl(C-B)$. Then, since we also have that $\Pl(A)
< \Pl(C - B)$, we can apply A2 to get that $\Pl(A \union B) < \Pl(C -
B) \le \Pl(C)$. But this contradicts our assumption that $\Pl(C) \le
\Pl(A \union B)$. Thus, we must have $B \not <^* (A \union B)$.
\eprf

\rethm{thm:Ranked}
If $(W,\Pl)$ be a rational qualitative plausibility space, then there is
a default-equivalent plausibility space $(W,\Pl')$ such that $\Pl'$ is a
ranking.
\erethm

\prf
Let $<^*$ be the relation defined in Lemma~\ref{lem:<^*}. Define a
relation $\approx^*$ on sets such that $A \approx^* B$ if neither $A <^*
B$ nor $B <^* A$. Since $<^*$ is modular, standard (and
straightforward) arguments show that
$\approx^*$ is an equivalence relation. We construct a new
plausibility measure based on these two ordering. Let $\F/{\approx^*}$
be the set of equivalence classes of measurable sets. Let $\le^*$ be
the the total order on $\F/\approx^*$ induced by $<^*$. Let $\Pl'$ be
a plausibility measure on $\F$ whose range is $\F/\approx^*$, defined
so that $\Pl'(A) = [A]$ where $[A]$ is the equivalence class
containing $A$. We
have to show that $\Pl'$ is a plausibility measure. Assume that $A
\subseteq B$. Then $\Pl(A) \le \Pl(B)$ and clearly $B \not<^*
A$. Thus, either $A <^* B$ or $A \approx^* B$. We conclude that
$\Pl'(A) \le \Pl'(B)$, as desired. Since $\le^*$ is a
total order, $\Pl'$ satisfies A4$'$. Using
Lemma~\ref{lem:<^*}(d), we get that $\Pl'(A\union
B) = \max(\Pl'(A),\Pl'(B))$. Thus, $\Pl'$ also
satisfies A5$'$, and hence is a ranking.
Finally, we have to show that
$(W,\Pl')$ is default-equivalent to $(W,\Pl)$. Let $A$ and $B$ be
disjoint events. 
By Lemma~\ref{lem:<^*}, $A <^* B$ if and only
if $\Pl(A) < \Pl(B)$. Thus, we conclude that $\Pl'(A) < \Pl'(B)$ if
and only if $\Pl(A) < \Pl(B)$.
\eprf

We now prove Theorem~\ref{thm:Ranked}.  We actually prove some more
general results.  Consider the following two restrictions on
plausibility structures:
\begin{description}\denselist
 \item[{\rm R}] $\Pl_w$ is rational for all worlds $w$.
 \item[{\rm N}] $\Pl_w$ in normal for all worlds $w$.
\end{description}

\thm\label{thm:cond-complete-extensions}
Let $\phi \in \LCond$, let $\A$ be a subset of $\{$R, N$\}$, and
let $A$ be the corresponding subset of $\{$C5, C6$\}$. Then
$\phi$ is valid in subset of $\PLclass_c^{QPL}$ that satisfies $\A$ if and
only if $\phi$ is provable from \SysC $+ A$.
\ethm

\prf
Again, we focus on completeness.
We obtain completeness in each case by modifying the proof of
    \Tref{SysCcomplete}. We construct a canonical model as in that proof,
    checking consistency with the extended axiom system. The resulting
    structure has the property that $(\PL,w_V) \sat \phi$ if
    and only if $\phi \in V$. We just need to show that this structure also
    satisfies the corresponding semantic restrictions.

If C5 is included as
an axiom, then C5 is valid in $\PL$. Using
\Pref{pro:rational} we get that $\PL$ is rational.

If C6 is included as an axiom
and $\PL$ is not normal, then there is some world $w_V$ where
$\Pl_{w_V}(\intension{\True}_{(\sPL,w_V)}) =
\Pl_{w_V}(\intension{\False}_{(\sPL,w_V)})$. We get that $(\PL,w_V)
\sat \True \Cond \False$, which contradicts C6.
\eprf

\lem\label{lem:finite-conditional-structure}
Let $\phi \in \LCond$ and let $\A$ be a subset of $\{$R, N$\}$.
If $\phi$ is satisfiable in a structure satisfying $\A$, it is
satisfiable in a finite structure satisfying $\A$.
\elem
\prf
Let $\phi \in \LCond$ and let $\PL = (W,\Plass, \pi)$ be a structure
satisfying $\A$ such that $(\PL,w) \sat \phi$ for some $w \in W$. We
now construct a finite structure $\PL'$ that satisfies $\phi$. The key
idea is that when evaluating $\phi$ we only examine subformulas of
$\phi$. Thus, we are  interested in distinguishing only between worlds that
differ in the evaluation of some subformula of $\phi$.

We now make this argument precise. Let $\Sub(\phi)$ be the set of
subformulas of $\phi$. We partition  $W$ into equivalence
classes: For  $w \in W$, define $[w] = \{ w' \in W : \forall \psi
\in \Sub(\phi), (\PL,w) \sat \psi \mbox{ if and only if } (\PL,w') \sat
\psi \}$. Thus, $[w]$ contains all the worlds that are
indistinguishable, for the purpose of evaluating $\phi$, from $w$.
We now choose, arbitrarily, a {\em representative\/} world
$\hat{w} \in [w]$ for each equivalence class. These definitions extend
to sets of worlds: For $A \subseteq W$, define $[A] = \union_{w \in
A} [w]$, and $\hat{A} = \{ \hat{w} : w \in A \}$.

We construct $\PL'$ as follows. We set $W' = \hat{W}$.
Clearly $W'$ is finite, since there are only a finite number of
subformulas of $\phi$. Let $\pi'$ be
the restriction of $\pi$ to $W'$, and let $\Plass'(\hat{w}) (W'_{\hat{w}}, \Pl'_{\hat{w}})$, where $W'_{\hat{w}} =
\hat{W_{\hat{w}}}$, and $\Pl'_{\hat{w}}$ is defined so that
$\Pl'_{\hat{w}}(A') \le
\Pl_{\hat{w}}(B')$ if $\Pl_{\hat{w}}([A'] \inter W_{\hat{w}}) \le
\Pl_{\hat{w}}([B'] \inter W_{\hat{w}})$.

We need show that $\PL'$ is a qualitative plausibility structure. This
is easy to prove since $[\cdot]$ preserves subsets, unions, and
disjointness of sets: if $A' \subseteq B'$, then $[A'] \subseteq [B']$; $[A'
\union B'] = [A']\union [B']$; and $[A' \inter B'] = [A'] \inter [B']$.

Next, we need to show that if $(\PL,w) \sat \phi$, then
$(\PL',\hat{w}) \sat \phi$. In fact, we prove this property for every
subformula $\psi$ of $\phi$. The proof is by induction on the structure of
of $\psi$. The only case of interest is for conditional formulas
$\psi \Cond \xi \in \Sub(\phi)$. This case follows easily once we
notice that if the induction hypothesis holds for $\psi$, then
$[\intension{\psi}_{(\sPL',\hat{w})}] \inter W_{\hat{w}} =
\intension{\psi}_{(\sPL,\hat{w})} \inter W_{\hat{w}}$.

Finally, we have to show that if $\PL$ is rational or normal, then so
is $\PL'$. Again, this follows easily from the properties of $[\cdot]$.
\eprf

\lem\label{lem:kappa-default-iso}
Let $(W,\F,\Pl)$ be a
rational and normal qualitative
plausibility space such that $\F$ is finite.
Then there is a $\kappa$-ranking $\kappa$ on $W$ and a possibility
measure $\Poss$ on $W$ such that
for all disjoint $A, B \in \F$,
$\Pl(A) > \Pl(B)$ if and only if $\kappa(A) < \kappa(B)$ and
$\Pl(A) > \Pl(B)$ if and only if $\Poss(A) > \Poss(B)$.
\elem

\prf
By Theorem~\ref{thm:Ranked}, there is a ranked plausibility
space $(W,\F,\Pl')$ such that $\Pl'$ is default-equivalent to
$\Pl$. Since $\F$ is finite, the set $\{ d : \exists A \in \F, \Pl'(A)
= d \}$ is finite. Moreover, since $\Pl'$ is a ranking, this set is
totally ordered. Let $d_0 > d_1 > \ldots > d_n$ be an ordered
enumeration of this set of values.

 Let $\kappa$ be a $\kappa$-ranking such that
$\kappa(A) = k$ if $\Pl'(A) = d_k > \bot$ and $\kappa(A) = \infty$ if
$\Pl'(A) = \bot$. Similarly, let $\Poss$ be a possibility measure such that
$\Poss(A) = 1 - k/n$ if $\Pl'(A) = d_k > \bot$ and $\Poss(A) = 0$ if
$\Pl'(A) = \bot$. It easy to see that since $\Pl'$ is a ranking, we
get that $\kappa(A \union B) = \min(\kappa(A),\kappa(B))$ and that
$\Poss(A\union B) = \max(\Poss(A),\Poss(B))$. It is also easy to see
that both measures are equivalent to $\Pl'$ and thus
default-equivalent to $\Pl$.
\eprf

\rethm{thm:AX-all}
\begin{enumerate}\denselist
 \item[(a)] {\bf C}$+ \{$C6$\}$ is a sound and complete axiomatization
of $\PLclass_c^\epsilon$.
 \item[(b)] {\bf C} $+ \{$C5, C6$\}$ is a sound and complete
axiomatization of $\PLclass_c^\kappa$
and $\PLclass_c^{Poss}$.
\end{enumerate}
\erethm

\prf
Part (a) is an immediate corollary of
Theorems~\ref{thm:epsilon-equivalent}
and~\ref{thm:cond-complete-extensions}.

For Part (b), as usual, it is easy to verify soundness.
For completeness, it suffices to show that if $\phi$ is consistent
with \SysC$ + \{$C5,C6$\}$, then it is satisfiable in $\PLclass_c^\kappa$
and $\PLclass_c^{Poss}$.
Suppose that $\phi$ is consistent
with \SysC$ + \{$C5,C6$\}$. By
Theorem~\ref{thm:cond-complete-extensions}, $\phi$ is satisfiable in
a
rational and normal structure in $\PLclass_C^{QPL}$.
By
Lemma~\ref{lem:finite-conditional-structure}, we can assume that this
structure is finite. The result now follows from
Lemma~\ref{lem:kappa-default-iso}.
\eprf

\end{document}